\documentclass{article} 
\usepackage{iclr2027_conference,times}

\usepackage{amsmath,amsfonts,bm}

\def\eqref#1{equation~\ref{#1}}

\def\1{\bm{1}}

\DeclareMathAlphabet{\mathsfit}{\encodingdefault}{\sfdefault}{m}{sl}
\SetMathAlphabet{\mathsfit}{bold}{\encodingdefault}{\sfdefault}{bx}{n}

\usepackage{hyperref}
\usepackage{url}
\usepackage{booktabs}
\usepackage{array}
\usepackage{tabularx}
\usepackage{array}
\newcolumntype{C}{>{\centering\arraybackslash}X}
\usepackage{multirow}
\usepackage{adjustbox}
\usepackage{wrapfig}
\usepackage[table]{xcolor}
\definecolor{SectionGray}{RGB}{247,247,247}
\definecolor{LPBlue}{RGB}{240,247,252}
\definecolor{LPBlueDark}{RGB}{229,240,248}
\definecolor{FTPurple}{RGB}{248,243,251}
\definecolor{FTPurpleDark}{RGB}{240,231,246}

\title{SleepXA: Composing Pretrained Neural and Cardiac Foundation Models for Sleep Representation Learning}

\author{%
\begin{minipage}{0.94\textwidth}
\raggedright
{\normalfont\small\bfseries
Xiaoyu~Chen$^{1,3}$, Xiao~Gu$^{2}$, Chao~Zhang$^{3}$,
Wen~Wu$^{3}$, Shuaicong~Hu$^{4}$,
Qinfan~Xiao$^{3}$, Jie~Yang$^{5}$\\
Chen~Chen$^{1}$, Wei~Chen$^{5}$}\\[0.5em]
{\normalfont\small
$^{1}$Fudan University \quad $^{2}$University of Oxford \quad
$^{3}$Shanghai Artificial Intelligence Laboratory\\
$^{4}$The University of Hong Kong \quad
$^{5}$The University of Sydney}\\[0.5em]
{\normalfont\scriptsize
\texttt{xiaoyuchen23@m.fudan.edu.cn},
\texttt{xiao.gu@eng.ox.ac.uk},
\texttt{zhangchao1@pjlab.org.cn}\\
\texttt{husc@hku.hk},
\texttt{chenchen\_fd@fudan.edu.cn},
\texttt{wei.chenbme@sydney.edu.au}}
\end{minipage}%
}

\iclrfinalcopy

\begin{document}

\maketitle
\lhead{Preprint}

\begin{abstract}
Physiological signals recorded during sleep provide synchronized views of neural, cardiac, respiratory, and other bodily systems, while preserving modality-specific information. Existing EEG and ECG foundation models already capture rich but distinct physiological structure, motivating an alternative to training a multimodal sleep model from scratch: composing existing pretrained models. We introduce SleepXA, a framework that uses synchronized polysomnography (PSG) to compose independently pretrained neural and cardiac foundation models. SleepXA combines pretrained EEG and ECG encoders with lightweight encoders for additional PSG modalities, and learns shared and modality-specific representations through sleep-guided alignment. We pretrain SleepXA on 9,230 overnight PSG recordings from three large cohorts and evaluate it on 27 downstream tasks spanning sleep representation, clinical and disease prediction, and daytime physiological transfer. SleepXA ranks among the top two methods on 22 of 27 tasks. 
Controlled experiments further show consistent benefits across sensor configurations and EEG–ECG backbone choices, while pretrained initialization outperforms architecture-matched training from scratch and SleepXA surpasses alternative multimodal composition strategies. These results suggest that synchronized multimodal physiology can serve as a natural interface for composing existing physiological foundation models, yielding reusable representations that extend beyond sleep to clinical and daytime physiological settings.
\end{abstract}

\section{Introduction}
\label{sec:introduction}
\vspace{-5pt}

Sleep is a fundamental physiological process supporting cognition, metabolism, cardiovascular function, and mental well-being~\citep{multi2026sleep}. Clinical sleep assessment commonly relies on Polysomnography (PSG), which synchronously records multiple physiological signals, including
electroencephalography (EEG), electrooculography (EOG),
electromyography (EMG), electrocardiography (ECG), respiratory signals,
and oxygen saturation (SpO$_2$)~\citep{berry2017aasm}.
Together, these modalities provide complementary yet coupled views of neural,
cardiac, respiratory, and other physiological processes during sleep, making
PSG a natural substrate for learning representations relevant to both sleep
analysis and broader health assessment.

Recent sleep foundation models (FMs) have demonstrated the value of large-scale
multimodal PSG pretraining for transferable sleep representations
~\citep{thapa2026sleepfm,huang2026sleepgpt,
yuan2026sleep2vec,shuai2026osf,hou2026omni,park2026sleepmami}.
Most existing approaches, however, treat heterogeneous PSG modalities largely
as parallel inputs for joint encoding or fusion, and they are mostly trained from scratch. Pretraining such a sleep foundation model requires the availability of well-curated and paired multimodal signals, at a scale and diversity that are prohibitively difficult to obtain, raising concerns about the representational breadth of the resulting models. Moreover, these sleep FMs are primarily optimized for shared sleep related information across signals, often without proper mechanisms to optimize modality-specific information.

On the other hand, large-scale pretraining has advanced general physiological representation learning~\citep{gu2025foundation}, and has resulted in a series of state-of-the-art and open-sourced foundation models on the specific modalities belonging to PSG, such as cardiac~\citep{li2025electrocardiogram,gu2026cardiac} and neural~\citep{jiang2024large,wang2025cbramod} signals. 

This motivates a complementary alternative to sleep representation learning:
\textit {Can we reuse and compose existing pretrained modality-specific FMs to extract good representations of sleep physiology?}

However, such composition is non-trivial for the following three reasons. 

(i) First, existing neural and cardiac FMs are pretrained independently, and mostly on non-sleep datasets. They may produce representations that are not naturally aligned due to signal semantics, temporal scales, training objectives, etc., thus making direct concatenation or fusion suboptimal. 

(ii) Second, different modalities may contain shared yet complementary information. Naïvely suppressing them into a common modality-invariant representation space may remove useful modality-specific information that is not captured by other modalities. 

(iii) Third, the available modalities during sleep monitoring may vary substantially across settings, from full PSG to reduced clinical
montages and wearable devices, making it challenging to learn representations that remain robust under partial modality availability.

To address these challenges, we present \textbf{SleepXA}, a compositional
framework that integrates independently pretrained modality-specific foundation models for sleep representation learning. SleepXA combines pretrained EEG and ECG FMs, with lightweight encoders for
additional PSG modalities, and learns to separate shared \emph{common} representation from modality-\emph{specific} information, with three fundamental modules (common space alignment, modality-specific preservation, and common-specific disentanglement) as detailed in the Methods Section.

We study this compositional paradigm at scale by pretraining SleepXA on 9,230 overnight PSG recordings from three large cohorts and evaluating it on 27 downstream tasks across 15 datasets, spanning general sleep representation, clinical and longitudinal prediction, and cross-context physiological transfer beyond sleep. SleepXA ranks first on 17 tasks and within the top two on 22.

In particular, extensive experiments demonstrate that SleepXA is not merely a ``better sleep model'', but also preserves broader modality-specific physiological knowledge inherited from the original FMs, while extending their capabilities to respiratory, cardiac, longitudinal, and out-of-sleep tasks. We further examine how pretrained initialization, EEG and ECG backbone choice, multimodal integration, sensor availability, and data-efficient adaptation shape these capabilities.

\begin{figure*}[!btp]
\centering
\includegraphics[width=0.93\textwidth]{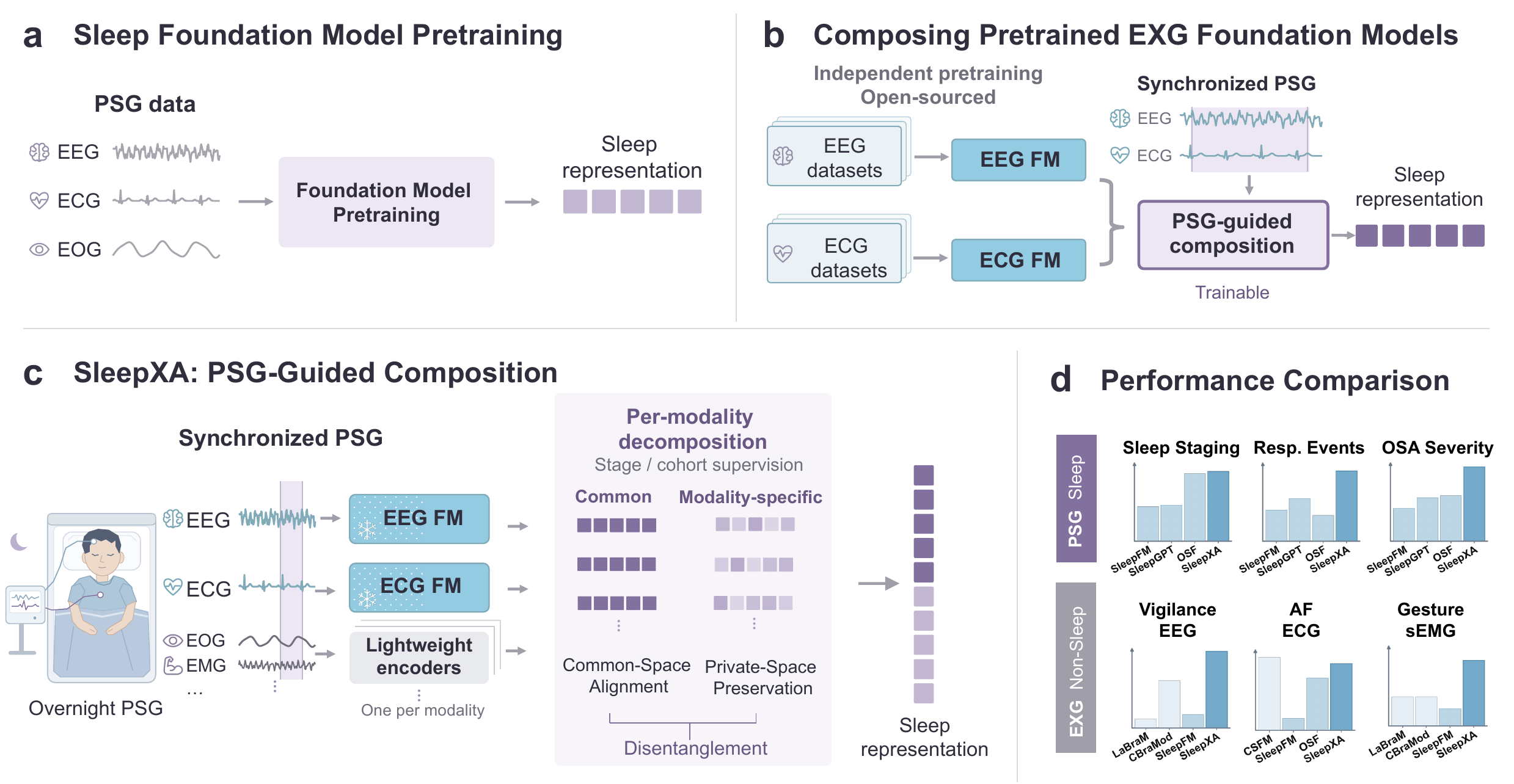}
\vspace{-10pt}
\caption{
\textbf{Overview of SleepXA.} \textbf{a.} Conventional sleep FMs are pretrained on polysomnography (PSG) from scratch. \textbf{b.} Independently pretrained, open-sourced EEG and ECG FMs can be composed for sleep representation learning. \textbf{c.} SleepXA aligns common representations, preserves modality-specific information, and reduces redundancy between modalities. \textbf{d.} Representative evaluation spans sleep and non-sleep tasks (normalized by panel, higher is better; details in Section \ref{sec:overall_performance}).
}
\label{fig:sleepxa_overview}
\vspace{-16pt}
\end{figure*}

\vspace{-10pt}
\section{Related Work}
\label{sec:related_work}
\vspace{-5pt}
\noindent\textbf{Sleep Foundation Models.}
Large-scale PSG pretraining has become an effective paradigm for transferable
sleep representation learning.
Existing approaches explore complementary directions, including cross-modal
contrastive learning in SleepFM~\citep{thapa2026sleepfm} and sleep2vec~\citep{yuan2026sleep2vec},
time-frequency and multi-objective pretraining in SleepGPT~\citep{huang2026sleepgpt},
robust pretraining and scaling across channel configurations in OSF~\citep{shuai2026osf}, hierarchical macro- and micro-structure modeling in SleepMaMi~\citep{park2026sleepmami}, and physiologically structured CNS-ANS modeling in
Omni-Sleep~\citep{hou2026omni}. These studies primarily learn multimodal physiological representations with a dedicated
sleep-pretraining pipeline.
SleepXA instead investigates whether transferable physiological knowledge already embedded
in independently pretrained EEG and ECG foundation models can be extracted and composed
through conditional shared-specific alignment to facilitate sleep representation learning.

\noindent\textbf{Physiological Foundation Models.}
Large-scale foundation models have also been developed for individual
physiological domains.
For neural signals, LaBraM~\citep{jiang2024large}, CBraMod~\citep{wang2025cbramod}, BrainOmni~\citep{xiao2026brainomni}, and REVE~\citep{el2026reve} learn transferable
representations from EEG or EEG/MEG recordings.
For cardiac signals, ECGFounder~\citep{li2025electrocardiogram} and CSFM~\citep{gu2026cardiac} learn reusable representations from
large-scale ECG or ECG/PPG corpora.
These models provide rich modality-specific physiology knowledge, but are pretrained independently and remain largely confined to their original physiological domains. Rather than simplistic late fusion to leverage these FMs in sleep PSG, SleepXA treats these as composable blocks and introduces a series of modules to align, preserve, and merge their representations for sleep representation learning.

\noindent\textbf{Multimodal Fusion and Alignment.}
Multimodal representations can be connected through direct feature fusion,
cross-attention, or contrastive alignment
~\citep{tsai2019multimodal,radford2021learning}.
Recent physiological studies further demonstrate cross-modal knowledge
transfer: AnyPPG uses a pretrained ECG encoder to guide PPG representation
learning~\citep{nie2026anyppg}, while MindMix aligns EEG with pretrained
acoustic representations~\citep{liu2026mindmix}.
Common-specific factorization has also been explored to separate shared and
modality-dependent information in multimodal learning
~\citep{qian2026decalign,meng2026dualbranch}.
However, it remains unclear how multiple modality-specific FMs can be aligned into a shared and reusable representation. This is particularly challenging in sleep PSG, where modality configurations vary substantially across datasets/deployment settings.

\vspace{-10pt}
\section{Method}
\vspace{-5pt}

\label{sec:method}

\vspace{-5pt}
\subsection{Motivation and Overview}
\label{sec:method_overview}
\vspace{-5pt}

SleepXA addresses the problem of composing physiological foundation models that were pretrained independently on different modalities and datasets. In our implementation, EEG and ECG are encoded by pretrained foundation models, while the remaining PSG modalities are represented using lightweight modality-specific encoders.

Following standard sleep-scoring conventions, a PSG recording can be segmented into consecutive 30-s epochs~\citep{berry2017aasm}. For epoch $i$ and modality
$m$, we define $\mathbf h_i^m = E_m(\mathbf x_i^m),$ where $\mathbf x_i^m$ denotes the 30-s input of modality $m$. 

For simplicity but without loss of generality, we use $E_m$ to denote the modality encoder in all cases: for EEG and ECG, $E_m$ is initialized from a pretrained FM (with the first several layers frozen, details in Appendix~\ref{app:implementation}), while for the remaining PSG modalities, $E_m$ denotes a lightweight encoder trained during SleepXA pretraining. Meanwhile, we denote their corresponding sleep-stage label as $y_i$ and cohort (or dataset) identity as $d_i$.

Directly combining these representations $\mathbf h_i^m$ is suboptimal for two reasons. First, independently pretrained encoders produce heterogeneous representation spaces that are not naturally aligned. Second, forcing all modalities into a single shared space may suppress physiological information that may be specific to individual modalities. 

SleepXA therefore decomposes each representation into
\vspace{-2pt}
\begin{equation}
\mathbf c_i^m = F^m_{\mathrm{com}}(\mathbf h_i^m),
\qquad
\mathbf s_i^m = F^m_{\mathrm{spec}}(\mathbf h_i^m),
\end{equation}
where $\mathbf h_i^m$, $\mathbf c_i^m$, and $\mathbf s_i^m$ denote its encoded
common and specific representations, from projector $F^m_{\mathrm{com}}$ and $F^m_{\mathrm{spec}}$ respectively.

\begin{figure*}[t]
    \centering
    \includegraphics[width=0.95\textwidth]{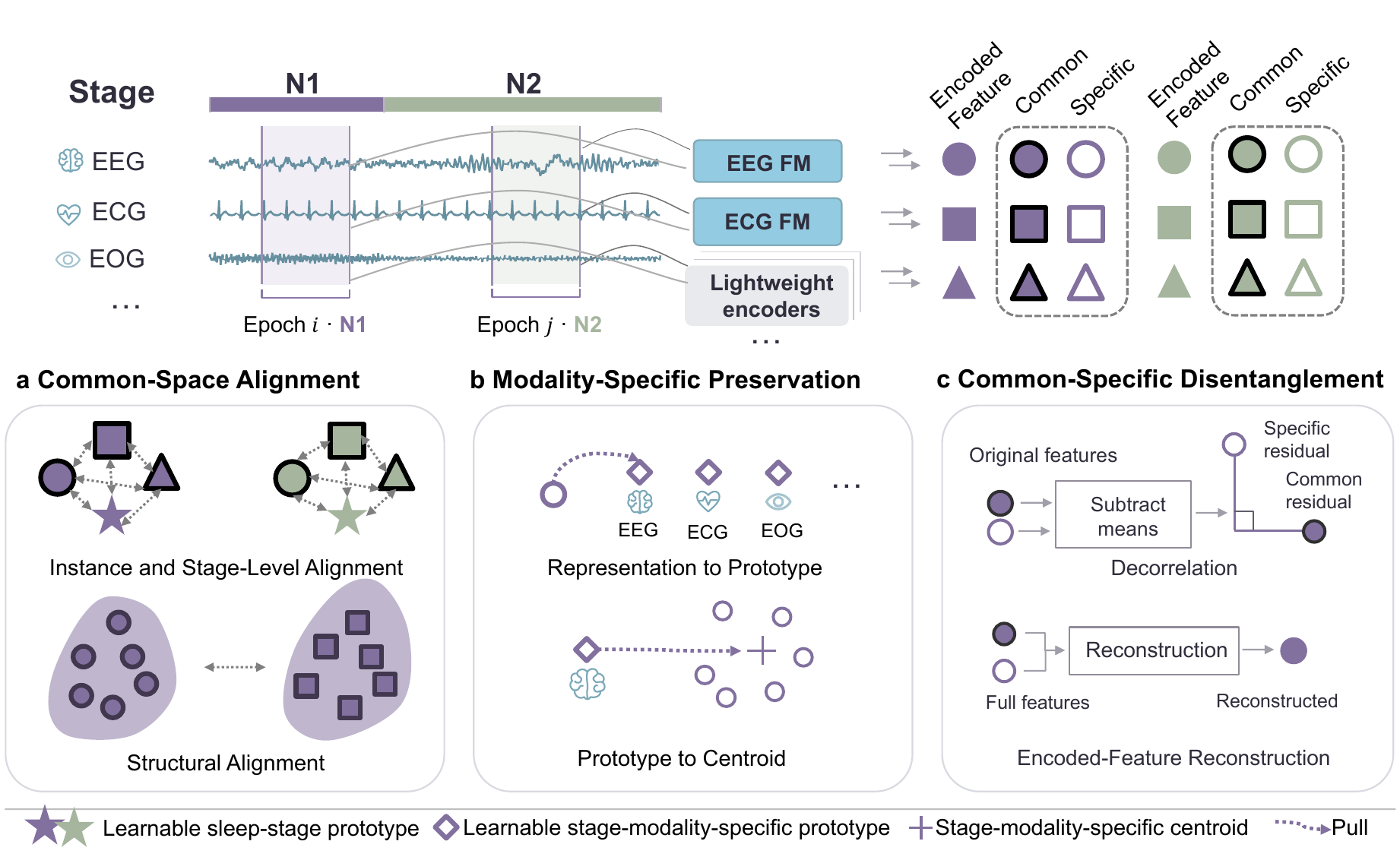}
    \vspace{-15pt}
    \caption{
\textbf{Sleep-guided cross-modality common-specific learning in SleepXA.} \textbf{a.} Common-space alignment combines instance- and stage-level alignment with distribution structural alignment within each cohort. \textbf{b.} Specific preservation matches each feature to its corresponding same-stage-modality prototypes while matching prototypes to their corresponding centroids. \textbf{c.} Disentanglement 
reduces redundancy between the two
components while retaining complementary physiology.
}
    \label{fig:sleepxa_alignment}
    \vspace{-15pt}
\end{figure*}

Below, we introduce three key modules, common-space alignment, modality-specific preservation, and common-specific disentanglement, as shown in \figureautorefname{}~\ref{fig:sleepxa_alignment}, to optimize the learning of $\mathbf c_i^m$ and $\mathbf s_i^m$.

\vspace{-10pt}
\subsection{Common-Space Alignment}
\vspace{-8pt}

\label{sec:common_alignment}

The first challenge is to align representations from heterogeneous modality encoders. Because PSG modalities are synchronously recorded, signals within the same 30-s epoch provide different physiological views of the same underlying state. We use this correspondence to organize the common representations at complementary levels, while discouraging them from encoding cohort identity.

\textbf{Instance and sleep-stage alignment.} We first align common representations $\mathbf c_i^m$ at two hierarchical levels, as indicated in \equationautorefname{}~\ref{eq:sleep_guided_alignment}. At the instance level, synchronized modality pairs {} are pulled together in a shared representation space. At the semantic level, representations from the same sleep stage $y_i$ are forced to be aligned to corresponding modality-invariant stage prototypes $\mathbf p_{y_i}$, which are learnable, similar to \cite{cui2021parametric}.
{Both objectives are implemented using contrastive learning, as detailed in Appendix~\ref{app:common_alignment}.}

\vspace{-6pt}
\begin{equation}
\vspace{-6pt}
\small
\mathcal L_{\mathrm{sem}}
=
\underbrace{
\mathcal L_{\mathrm{inst}}({\mathbf c_i^m})
}_{\text{instance level}}
+
\alpha_{\mathrm{stage}}
\underbrace{
\mathcal L_{\mathrm{stage}}({\mathbf c_i^m},{\mathbf p_{y_i}})
}_{\text{sleep-stage level}} .
\label{eq:sleep_guided_alignment}
\end{equation}

\textbf{Structural alignment.} The above semantic alignment introduces alignment at both individual and stage level. However, at the global distribution level, it is also expected that for the same sleep stage, the distribution of different modalities, such as $\{\mathbf c_i^m : m=\mathrm{EEG}, y_i=N_1\}$ and $\{\mathbf c_i^m : m=\mathrm{ECG}, y_i=N_1\}$, are expected to exhibit similar feature distributions. This is analogous to previous research pursuing distribution alignment across semantically similar categories~\citep{wang2023few,sun2016deep}. To realize this, we perform conditional distribution alignment across modalities within each cohort-stage group. 

\begin{equation}
\vspace{-4pt}
\small
\mathcal L_{\mathrm{struc}}
=
\mathbb E_{d,y}
\sum_{\substack{
\{m,n\}\subseteq\mathcal M\\
m\neq n
}}
D_{\mathrm{struct}}
\left(
\{\mathbf c_i^m\}_{d,y},
\{\mathbf c_i^n\}_{d,y}
\right).
\label{eq:conditional_alignment}
\end{equation}
where $d$ and $y$ denote cohort and sleep stage, respectively, and $D_{\mathrm{struct}}$ measures the discrepancy between the conditional representation structures of two modalities. Please refer to Appendix~\ref{app:common_alignment} for implementation details.

The resulting common-space objective is
$
\scriptsize
\mathcal L_{\mathrm{com}}
=
\mathcal L_{\mathrm{sem}}
+
\mathcal L_{\mathrm{struc}} .
$

\vspace{-10pt}
\subsection{Modality-Specific Preservation}
\vspace{-5pt}
\label{sec:specific_preservation}
While the common branch $F^m_{com}$ promotes cross-modal consistency, the specific branch $F^m_{spec}$ 
is supposed to retain physiological information unique to each individual modality. We achieve this by encouraging modality-specific representations to remain distinguishable within each sleep stage. This is different from naively enforcing global modality discriminability, which may conflate modality-specific variation with differences arising from sleep stage.

To implement this, we introduce learnable prototypes $\mathbf q_y^m$ for each modality $m$ and sleep stage $y$. For a specific representation $\mathbf s_i^m$ with stage label $y_i$, we encourage it to be closer to the prototype of its source modality than to those of other modalities under the same sleep stage. We further anchor each prototype to the corresponding modality distribution using its stage-conditioned centroid. The resulting objective is
\vspace{-3pt}
\begin{equation}
\small
\mathcal L_{\mathrm{spec}}
=
\underbrace{
\mathcal L_{\mathrm{proto}}^{\rightarrow}
\left(
\{\mathbf s_i^m,y_i\},
\{\mathbf q_y^m\}
\right)
}_{\text{representation-to-prototype}}
+
\underbrace{
\mathcal L_{\mathrm{proto}}^{\leftarrow}
\left(
\{\mathbf q_y^m\},
\{\bar{\mathbf s}_{d,y}^m\}
\right)
}_{\text{prototype-to-centroid}}.
\label{eq:specific_loss}
\end{equation}
\vspace{-3pt}
We implement both terms using contrastive objectives. Here, $\bar{\mathbf s}_{d,y}^m$ denotes the centroid of modality $m$ within cohort $d$ and sleep stage $y$. Because not all modalities are available in every cohort, we construct these centroids separately within each cohort-stage condition, using only the modalities observed in that cohort. The first term preserves modality identity within each sleep stage, while the second keeps the learned prototypes representative of their corresponding modality distributions. Detailed formulations are provided in Appendix~\ref{app:common_alignment}.

\vspace{-10pt}
\subsection{Common-Specific Disentanglement}
\vspace{-5pt}

\label{sec:conditional_disentanglement}

Although the common and specific branches are optimized for different purposes, they may still encode redundant information. For example, both $\mathbf c_i^m$ and $\mathbf s_i^m$ may vary with sleep stage, such as between $N_2$ and REM, and this is not necessarily undesirable. The problem is when they continue to encode the same variation even within the same sleep stage and cohort. We therefore avoid enforcing global orthogonality, which could remove meaningful stage- or cohort-related information shared by both branches.

We therefore measure redundancy relative to each cohort-stage condition (refer to Appendix~\ref{app:major_ablation} for an ablation
of this constraint). Specifically, we remove the variation explained by the corresponding context and define
\begin{equation}
\small
\Delta \mathbf c_i^m
=
\mathbf c_i^m-\bar{\mathbf c}_{d,y}^m,
\qquad
\Delta \mathbf s_i^m
=
\mathbf s_i^m-\bar{\mathbf s}_{d,y}^m,
\end{equation}
We then decorrelate only the remaining residual variation:

\begin{equation}
\small
\mathcal L_{\mathrm{dis}}
=
\underbrace{
\mathbb E_{i,m}
\left[
\cos^2
\left(
\Delta\mathbf c_i^m,
\Delta\mathbf s_i^m
\right)
\right]
}_{\text{conditional decorrelation}}
+
\lambda_{\mathrm{rec}}
\underbrace{
\mathbb E_{i,m}
\left[
\left\|
\mathbf h_i^m
-
R_m([\mathbf c_i^m;\mathbf s_i^m])
\right\|_2^2
\right]
}_{\text{representation reconstruction}} .
\label{eq:disentanglement}
\end{equation}

Here, $\cos(\cdot,\cdot)$ denotes cosine similarity between the two residual representations, while the latter reconstruction, implemented by a lightweight MLP $R_m$, prevents the common-specific decomposition from discarding information contained in the original encoded representation. Together, these objectives reduce redundancy between the common and specific branches without removing meaningful variation associated with sleep stage or cohort.

\vspace{-10pt}
\subsection{Overall Objective and Downstream Representation}
\vspace{-8pt}
\paragraph{Overall training objective.}
SleepXA combines three complementary modules introduced above. The overall training objective is realized as follows, 

\begin{equation}
\small
\mathcal L_{\mathrm{total}}=\mathcal L_{\mathrm{com}}
+
\lambda_{\mathrm{spec}}\mathcal L_{\mathrm{spec}}
+
\lambda_{\mathrm{dis}}\mathcal L_{\mathrm{dis}}.
\label{eq:total_loss}
\end{equation}

Further details of hyperparameter settings are available in Appendix~\ref{app:common_alignment}. 

\vspace{-6pt}
\paragraph{Downstream representation.}
For downstream tasks, the common and modality-specific representations from the
available modalities are combined to form the task representation, with the
representation pathway adapted according to the downstream evaluation setting. For linear probing, we
average the available common features and concatenate the result with
the modality-specific features. For fine-tuning, task-adaptive attention
separately fuses the common and specific features across available modalities.

\vspace{-12pt}
\section{Experiments and Results}
\vspace{-5pt}
\label{sec:experiments}
\vspace{-5pt}
\subsection{Experimental Setup}
\label{sec:experimental_setup}
\vspace{-5pt}

\textbf{Pre-training and downstream data.}
We perform multimodal alignment pre-training on 9,230 overnight PSG recordings from SHHS, MESA, and CFS~\citep{quan1997sleep,chen2015racial,redline1995familial}.
Downstream evaluation spans 27 tasks across held-out Test SHHS1/2 and seven external sleep cohorts (APPLES, WSC, SleepEDF20, HMC, ISRUC-1, MASS-SS2, and CAP)~\citep{quan2011association,young2009burden,kemp2000analysis,alvarez2021inter,khalighi2016isruc,o2014montreal,terzano2001atlas}, together with seven non-sleep physiological datasets covering EEG, ECG, EOG, and EMG~\citep{mumtaz2017electroencephalogram,miltiadous2023dataset,zyma2019electroencephalograms,tangermann2012review,clifford2017af,zuniga2025surfaceemg,zheng2017multimodal}.
We evaluate SleepXA across these 15 datasets on 27 downstream tasks in three groups: \textit{general sleep representation}, \textit{clinical and longitudinal prediction}, and \textit{sleep-to-non-sleep physiological transfer}.
Complete dataset statistics, split construction, leakage control, and task definitions are provided in Appendix~\ref{app:datasets}.

\vspace{-10pt}

\paragraph{Evaluation and baselines.}
Foundation models are evaluated under linear probing (LP) and full fine-tuning
(FT), 
using cross-subject splits whenever participant
identifiers are available. 

We compare SleepXA against three groups of baselines: task-specific supervised models; physiological foundation models, including LaBraM~\citep{jiang2024large} and CBraMod~\citep{wang2025cbramod} for EEG and CSFM~\citep{gu2026cardiac} for ECG; and sleep foundation models, with SleepFM~\citep{thapa2026sleepfm},
SleepGPT~\citep{huang2026sleepgpt}, and OSF~\citep{shuai2026osf}
as the main baselines.
Unless otherwise specified, SleepXA uses LaBraM and CSFM as its pretrained EEG and ECG encoders, respectively.
All methods use the same splits and metrics, with supported input modalities. Detailed optimization settings and task-specific evaluation protocols are provided in Appendices~\ref{app:implementation} and~\ref{app:baselines}.

\vspace{-10pt}
\subsection{Performance across Downstream Tasks}
\label{sec:overall_performance}
\vspace{-5pt}

\begin{figure*}[t]
    \centering
    \includegraphics[width=\textwidth]{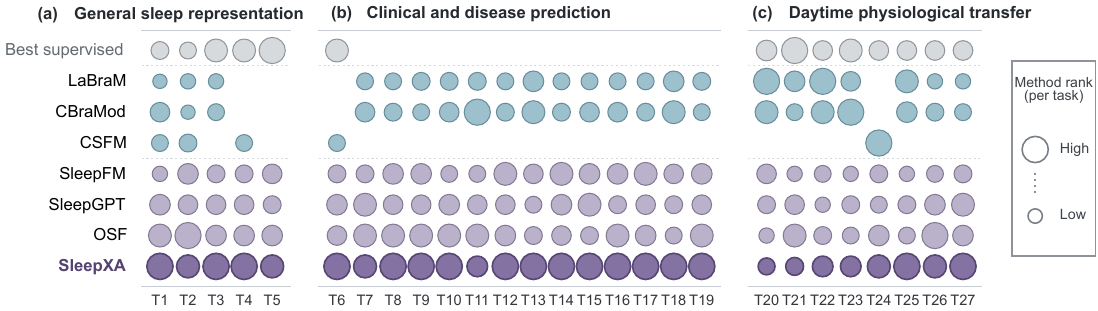}
    \vspace{-20pt}
    \caption{
    \textbf{Cross-task performance landscape.}
    Results span general sleep representation, clinical and longitudinal
    prediction, and daytime physiological transfer.
    Larger bubbles indicate better task-wise ranks.
    Gray, blue, lavender, and dark-purple denote supervised references,
    physiological foundation models, sleep foundation models, and SleepXA,
    respectively. Task definitions for T1--T27 are provided in     Appendix~\ref{app:datasets}.  
    }
    \label{fig:cross_task_summary}
     \vspace{-15pt}
\end{figure*}

Figure~\ref{fig:cross_task_summary} summarizes performance across the 27-task benchmark. To emphasize pretrained representation quality, we use an LP-first ranking protocol, with full details provided in Appendix~A.3. \textbf{SleepXA ranks first on 17 of 27 tasks and within the top two on 22},
demonstrating competitive performance across all three evaluation groups.
The advantage is particularly consistent for general sleep representation and
clinical prediction,
while meaningful transfer is also retained on cross-context physiological
tasks involving EEG, ECG, and multimodal inputs.
Together, these results highlight the strength and broad transferability of
SleepXA representations across diverse downstream settings.

\begin{figure*}[t]
    \centering
    \includegraphics[width=\textwidth]{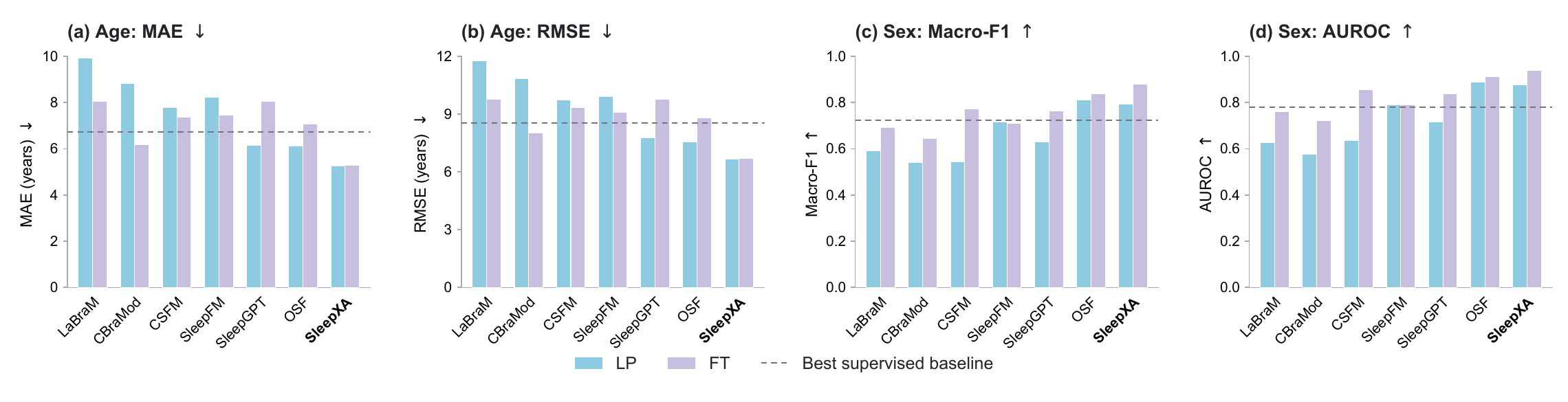}
    \vspace{-25pt}
    \caption{
    \textbf{Age and sex prediction.}
    Performance of foundation models under LP and FT.
    Age prediction is evaluated by MAE and RMSE ($\downarrow$), while sex prediction is
    evaluated by Macro-F1 and AUROC ($\uparrow$).
    Dashed lines denote the best supervised baselines.
    }
    \label{fig:age_sex_prediction}
    \vspace{-15pt}
\end{figure*}

\vspace{-5pt}
\paragraph{General sleep representation.}We first evaluate whether SleepXA captures general sleep-related physiology at multiple prediction granularity levels. At the subject level, SleepXA achieves the strongest age-estimation performance under both LP and FT, while achieving the best sex-classification performance after FT (Fig.~\ref{fig:age_sex_prediction}, Table ~\ref{tab:age_sex_results}).
We next evaluate epoch-level sleep staging as a direct test of cross-cohort transfer. 
On Test SHHS1/2 and the two external cohorts APPLES and WSC, SleepXA achieves the best performance among the evaluated foundation models under LP, reaching accuracy scores of 84.07\%, 77.63\%, and 85.16\%, respectively (Table~\ref{tab:sleep_staging_results1}).
FT further improves performance on Test SHHS1/2 and WSC while maintaining comparable performance on APPLES, indicating that the representation is both directly transferable and adaptable to downstream supervision. Further sleep staging evaluations are provided in Appendix~\ref{app:detailed_downstream_results}.
Finally, SleepXA also shows strong transfer to fine-grained respiratory-event (Table~\ref{tab:resp_event_results}) and spindle detection (Appendix~\ref{absec:spindle_detection}), suggesting that its representation captures broader sleep physiology beyond sleep-stage classification.

\begin{table}[t]
\centering
\caption{
\textbf{Sleep staging on three representative cohorts.} Acc and MF1 are percentages; all metrics are higher-is-better. \textbf{Bold} and \underline{underlined} values indicate the \textbf{best} and \underline{second-best} results.
}
\label{tab:sleep_staging_results1}

\begingroup
\footnotesize
\setlength{\tabcolsep}{1.7pt}
\renewcommand{\arraystretch}{0.92}

\begin{tabularx}{0.97\linewidth}{
@{}lc*{9}{C}@{}
}
\toprule

\multirow{2}{*}{\textbf{Model}}
& \multirow{2}{*}{\textbf{Channels}}
& \multicolumn{3}{c}{\textbf{Test SHHS1/2}}
& \multicolumn{3}{c}{\textbf{APPLES}}
& \multicolumn{3}{c}{\textbf{WSC}} \\

\cmidrule(lr){3-5}
\cmidrule(lr){6-8}
\cmidrule(l){9-11}

& & Acc. & MF1 & $\kappa$
  & Acc. & MF1 & $\kappa$
  & Acc. & MF1 & $\kappa$ \\

\midrule
\rowcolor{SectionGray}
\multicolumn{11}{@{}l}{
\textit{Task-specific supervised models}
} \\
\addlinespace[0.5pt]

DeepSleepNet{\scriptsize~\citep{supratak2017deepsleepnet}} & E
& 80.10 & 72.66 & 0.728
& 75.12 & 67.87 & 0.658
& 81.89 & 74.70 & 0.732 \\

SleePyCo{\scriptsize~\citep{lee2024sleepyco}} & E
& 82.12 & 74.62 & 0.753
& 76.57 & 65.92 & 0.667
& 83.23 & 75.78 & 0.748 \\

U-Time{\scriptsize~\citep{perslev2019u}} & E
& 79.10 & 70.58 & 0.712
& 76.26 & 67.80 & 0.665
& 83.11 & 75.26 & 0.745 \\

AttnSleep{\scriptsize~\citep{eldele2021attention}} & E
& 74.98 & 63.59 & 0.650
& 73.42 & 58.00 & 0.616
& 80.48 & 72.82 & 0.712 \\

XSleepNet{\scriptsize~\citep{phan2021xsleepnet}} & E/O/M
& 84.41 & \underline{77.38} & \underline{0.783}
& 76.81 & 70.20 & 0.678
& 84.70 & 77.47 & 0.769 \\

ASTGSleep{\scriptsize~\citep{chen2025astgsleep}} & Full
& 83.81 & 77.08 & 0.777
& 76.88 & 70.25 & \underline{0.690}
& \textbf{85.97} & \textbf{78.25} & \underline{0.782} \\

XSleepFusion{\scriptsize~\citep{hu2025xsleepfusion}} & Full
& 82.44 & 73.95 & 0.756
& 75.99 & 68.09 & 0.665
& 83.01 & 75.98 & 0.747 \\
SleepXViT{\scriptsize~\citep{lee2025explainable}} & Full
& 83.89 & 71.54 & 0.767
& 76.41 & 65.12 & 0.662
& 83.40 & 73.49 & 0.741 \\

\midrule
\rowcolor{LPBlue}
\multicolumn{11}{@{}l}{
\textit{Foundation models: linear probing (LP)}
} \\
\addlinespace[0.5pt]

LaBraM{\scriptsize~\citep{jiang2024large}} & E
& 74.65 & 61.57 & 0.639
& 67.40 & 59.85 & 0.556
& 72.35 & 60.40 & 0.577 \\

CBraMod{\scriptsize~\citep{wang2025cbramod}} & E
& 76.52 & 64.44 & 0.667
& 69.57 & 63.62 & 0.588
& 73.18 & 62.01 & 0.594 \\

SleepFM{\scriptsize~\citep{thapa2026sleepfm}} & Full
& 75.22 & 68.05 & 0.667
& 70.89 & 62.70 & 0.600
& 77.40 & 67.97 & 0.658 \\

SleepGPT{\scriptsize~\citep{huang2026sleepgpt}} & Full
& 77.91 & 69.75 & 0.697
& 72.24 & 63.92 & 0.615
& 79.51 & 68.35 & 0.681 \\

SleepMaMi{\scriptsize~\citep{park2026sleepmami}} & Full
& 80.28 & 69.92 & 0.721
& 75.28 & 65.69 & 0.648
& 81.33 & 71.51 & 0.708 \\

OSF{\scriptsize~\citep{shuai2026osf}} & Full
& 83.14 & 74.42 & 0.764
& 75.50 & 67.30 & 0.660
& 83.52 & 75.61 & \textbf{0.794} \\

\addlinespace[2pt]
\rowcolor{LPBlue}
\textbf{SleepXA (Ours)} & Full
& 84.07 & 76.24 & 0.780
& \underline{77.63} & \textbf{71.13} & \textbf{0.692}
& 85.16 & 76.09 & 0.765 \\

\midrule
\rowcolor{FTPurple}
\multicolumn{11}{@{}l}{
\textit{Foundation models: full fine-tuning (FT)}
} \\
\addlinespace[0.5pt]

LaBraM{\scriptsize~\citep{jiang2024large}} & E
& 83.43 & 72.97 & 0.765
& 75.56 & 67.57 & 0.660
& 79.73 & 70.04 & 0.692 \\

CBraMod{\scriptsize~\citep{wang2025cbramod}} & E
& 83.91 & 74.76 & 0.774
& 76.60 & 68.05 & 0.671
& 81.36 & 71.03 & 0.712 \\

SleepFM{\scriptsize~\citep{thapa2026sleepfm}} & Full
& 74.51 & 67.73 & 0.658
& 72.09 & 62.95 & 0.610
& 68.73 & 63.79 & 0.562 \\

SleepGPT{\scriptsize~\citep{huang2026sleepgpt}} & Full
& 82.13 & 74.37 & 0.753
& 73.06 & 66.51 & 0.632
& 82.46 & 71.54 & 0.722 \\

SleepMaMi{\scriptsize~\citep{park2026sleepmami}} & Full
& \underline{84.64} & 75.13 & 0.781
& 77.47 & 67.91 & 0.681
& 85.16 & 77.09 & 0.771 \\

OSF{\scriptsize~\citep{shuai2026osf}} & Full
& 84.09 & 75.58 & 0.782
& 76.72 & 68.20 & 0.670
& 83.18 & 74.97 & 0.742 \\

\addlinespace[2pt]
\rowcolor{FTPurple}
\textbf{SleepXA (Ours)} & Full
& \textbf{87.04} & \textbf{78.71} & \textbf{0.815}
& \textbf{77.84} & \underline{70.51} & \underline{0.690}
& \underline{85.95} & \underline{77.61} & 0.780 \\

\bottomrule
\end{tabularx}

\par\vspace{3pt}
\begin{minipage}{\linewidth}
\scriptsize

{E: EEG; E/O/M: EEG+EOG+EMG; Full: full experimental PSG configuration.}
\end{minipage}

\endgroup
\vspace{-15pt}
\end{table}

\vspace{-10pt}
\paragraph{Clinical and longitudinal prediction.}
We next assess whether SleepXA demonstrates transferable capability to predict clinically meaningful subject-level outcomes. 
For sleep-disordered breathing, we evaluate both continuous AHI estimation and four-level OSA severity classification on Test SHHS1/2 and WSC. As shown in Table~\ref{tab:osa_severity_results},
SleepXA shows strong cross-cohort transfer already under LP and further improves with FT, reaching MAEs of 4.200 on Test SHHS1/2 and 3.120 on WSC. 
These results indicate that the learned representation supports both continuous apnea-burden estimation and clinically defined OSA severity assessment.
More broadly, across 12 frozen-representation clinical tasks, including six current disease diagnoses, two longitudinal sleep outcomes, and four five-year cardiovascular outcomes, SleepXA achieves the highest mean AUROC among the evaluated foundation models on 11 tasks, including five of six current diagnoses and all six longitudinal outcomes (Tables~\ref{tab:current_disease_results} and
\ref{tab:longitudinal_disease_results}).
Given the limited clinical sample size, these findings provide preliminary evidence of transferability to clinical and longitudinal prediction and warrant further validation in larger cohorts.
More details are in Appendices~\ref{app:clinical_results} and \ref{app:clinical_prediction}.

\begin{figure*}[t]
    \centering
    \includegraphics[
        width=\textwidth,
        trim={0 2pt 0 0},
        clip
    ]{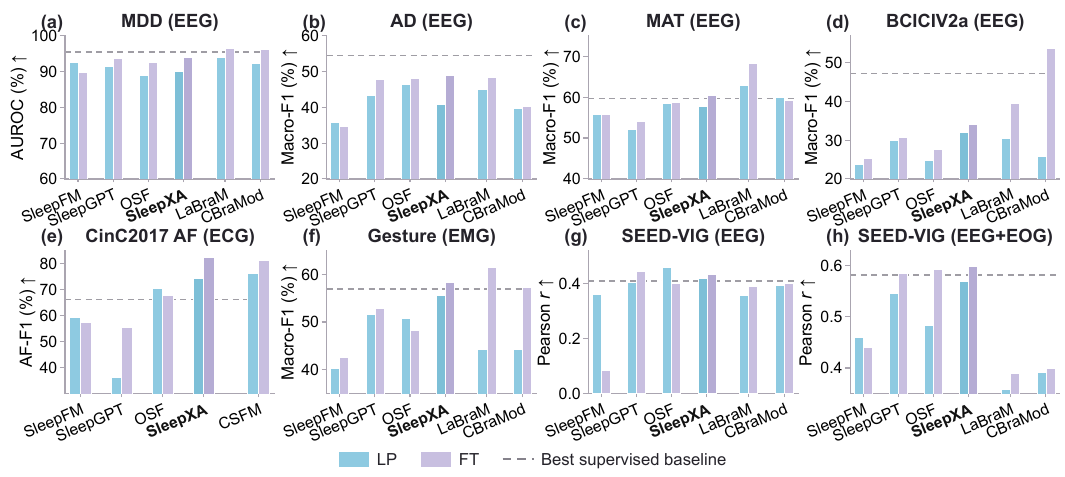}
    \vspace{-20pt}
    \caption{
    \textbf{Cross-context physiological transfer beyond nocturnal sleep.}
    Performance across eight downstream settings involving EEG, ECG, EOG,
    and surface EMG.
    Foundation models are evaluated under LP and FT, and dashed lines denote the best supervised
    baseline for each setting. 
    }
    \vspace{-8pt}
    \label{fig:daytime_transfer}
\end{figure*}

\vspace{-10pt}
\paragraph{Daytime physiological transfer.}
We further examine whether representations learned from nocturnal PSG transfer beyond the sleep setting. 
As shown in Fig.~\ref{fig:daytime_transfer}, these results highlight an underexplored property of PSG-guided physiology FM composition: \textit{nocturnal PSG appears to facilitate learning whole-body physiological structure, yielding knowledge that remains reusable beyond sleep itself.}
(i) First, SleepXA supports decoding brain states that reflect health conditions and physiological function beyond sleep. It transfers to multiple tasks, such as Major depressive disorder (MDD) and Alzheimer's disease (AD) (Fig.~\ref{fig:daytime_transfer} a-b). (ii) Second, SleepXA remains competitive even when the downstream task is semantically distant from sleep.  
On EEG based mental arithmetic recognition (MAT) and motor-imagery decoding, as well as ECG-based atrial-fibrillation (AF) recognition, SleepXA ranks consistently among the top two sleep FMs, demonstrating the effective use of the pretrained modality-specific encoders beyond sleep-related semantics (Fig.~\ref{fig:daytime_transfer} c-e). (iii) Third, SleepXA supports both cross-modality transfer and multimodal integration. 
It performs competitively on EMG-based gesture recognition and EEG-only vigilance estimation, and achieves the best linear-probing performance when EEG and EOG are jointly used, supporting the utility of the learned representations when modalities are used individually or combined (Fig.~\ref{fig:daytime_transfer} f-h).
These results support the value of PSG-based multimodal alignment for organizing complementary modality-specific information.
See Appendix~\ref{app:daytime_transfer_discussion} for further discussion.

\begin{figure}[!t]
\centering

\begin{minipage}[t]{0.49\columnwidth}
    \centering
    \includegraphics[width=\linewidth]{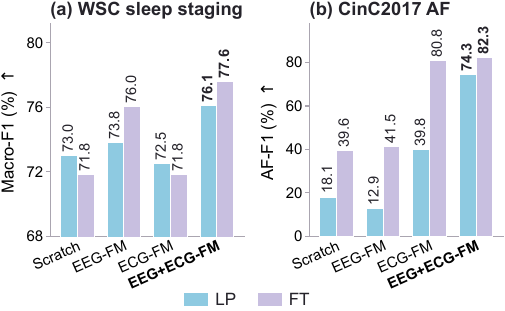}
    \vspace{-25pt}
    \caption{
    \textbf{Effect of physiological foundation-model initialization.}
    Comparison on
    (a) WSC sleep staging and
    (b) CinC2017 AF recognition.
    }
    \label{fig:fm_initialization}
\end{minipage}
\hfill
\begin{minipage}[t]{0.49\columnwidth}
    \centering
    \includegraphics[width=\linewidth]{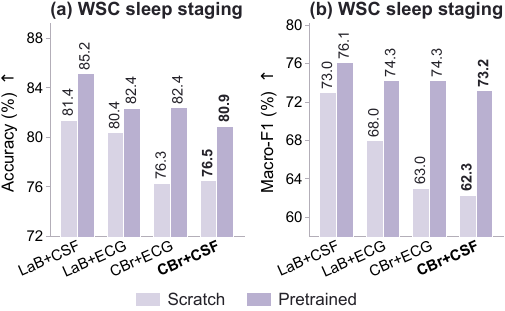}
    \vspace{-25pt}
    \caption{
    \textbf{Scratch vs pretrained initialization across FM combinations.} LaB:LaBraM, CBr:CBraMod, CSF:CSFM, ECG:ECGFounder.
 }
    \label{fig:backbone_robustness}
\end{minipage}

\vspace{-8pt}
\end{figure}

\vspace{-5pt}
\subsection{Analysis of Model Composition}
\label{sec:analysis}
\vspace{-5pt}
\paragraph{Effect of physiological foundation-model initialization.}

To assess the contribution of pretrained physiological knowledge during composition, we investigate different EEG/ECG encoder initialization strategies while keeping all other settings fixed. 
As shown in Fig.~\ref{fig:fm_initialization}, joint EEG+ECG foundation-model
initialization performs best on both WSC sleep staging (full PSG) and CinC2017 AF
recognition (ECG), 

supporting the benefit of composing complementary pretrained neural
and cardiac knowledge through synchronized PSG alignment.
Complete results are provided in
Appendix~\ref{app:fm_initialization}.

\vspace{-10pt}
\paragraph{Robustness to physiological foundation-model backbones.} To test the FM-agnostic of our composition, we test all $2 \times 2$ combinations of two EEG foundation models (LaBraM and CBraMod) and two ECG foundation models (CSFM and ECGFounder~\citep{li2025electrocardiogram}). Across all four pairs, pretrained initialization outperforms matched scratch training by 1.98-6.12 points in accuracy on WSC sleep staging (Fig.~\ref{fig:backbone_robustness}; Appendix~\ref{app:backbone_robustness}),
showing that the gains are not specific to the default LaBraM+CSFM
configuration.

\begin{figure*}[t]
    \centering
    \includegraphics[
        width=\textwidth,
        trim={0 0 0 0},
        clip
    ]{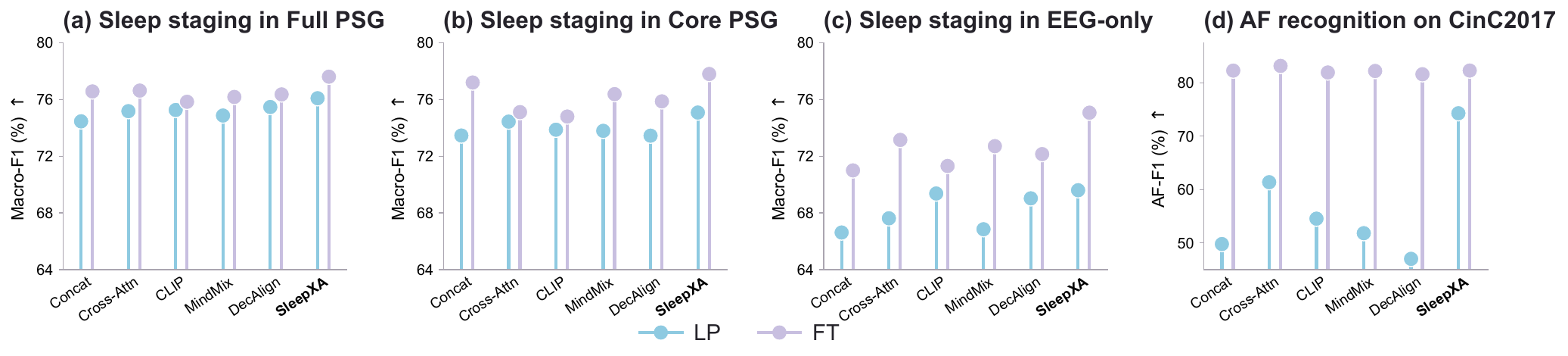}
    \vspace{-25pt}
\caption{
\textbf{Comparison of multimodal integration strategies.}
WSC sleep staging under (a) Full, (b) Core, and (c) EEG-only inputs,
and (d) CinC2017 AF recognition under LP and FT.
}
    \label{fig:integration_comparison}
    \vspace{-5pt}
\end{figure*}

\vspace{-10pt}
\paragraph{Comparison with multimodal integration strategies.}
We compare SleepXA with representative fusion and alignment alternatives in four configurations: WSC sleep staging with full-PSG, core-PSG (EEG+EOG+EMG), or EEG-only inputs, and ECG AF recognition on CinC2017. 
SleepXA performs best across all WSC settings and shows a clear LP advantage on CinC2017, while FT largely narrows the gap (Fig.~\ref{fig:integration_comparison}).
These results suggest that SleepXA's alignment improves the quality of frozen representations by preserving and organizing complementary pretrained modality-specific knowledge. 
More details are provided in Appendix~\ref{app:integration_comparison}.

\begin{figure*}[t]
    \centering
    \includegraphics[
        width=\textwidth,
        trim={0 0 0 0},
        clip
    ]{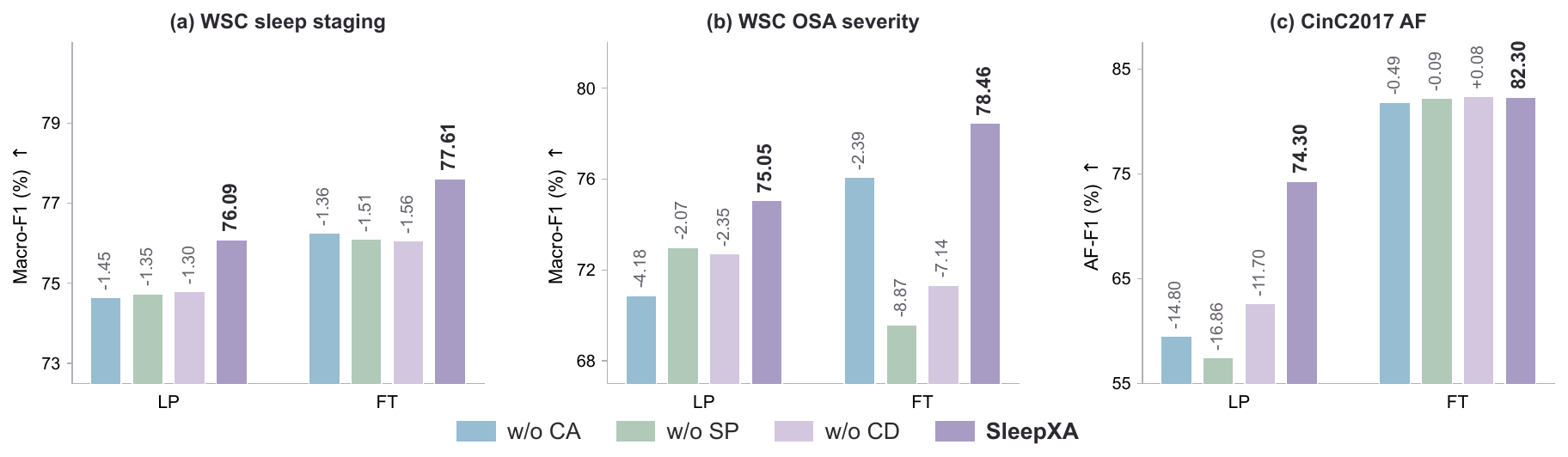}
    \vspace{-25pt}
\caption{
\textbf{Ablation of the major alignment objectives.}
Effects of removing CA, SP, or CD on
(a) WSC sleep staging,
(b) WSC OSA severity classification, and
(c) CinC2017 AF recognition.
Annotations report the change relative to the original SleepXA model under
the same LP/FT setting.
}
    \label{fig:major_ablation}
    \vspace{-20pt}
\end{figure*}

\vspace{-10pt}
\paragraph{Ablation of major alignment objectives.}
We independently remove condition-aware alignment (CA), stage-conditioned
specific preservation (SP), and condition-relative disentanglement (CD).
As shown in Fig.~\ref{fig:major_ablation}, removing any module degrades
WSC sleep staging and OSA severity prediction under both LP and FT. On CinC2017, alignment improves LP but adds little after FT, as FT may unlock AF-relevant knowledge already encoded in the pretrained ECG FM.
Together, these results demonstrate each module's complementary contribution, with their
larger effect under LP indicating that they primarily improve pretrained
representation quality
(Appendix~\ref{app:major_ablation}).

\vspace{-10pt}
\paragraph{Alignment with limited sleep stage labels.}
To evaluate SleepXA under conditions of limited sleep stage labels, we compare
four pretraining alignment settings: full labels, no stage guidance,
hybrid pseudo label guidance, and partial label guidance.
The latter two settings retained the expert SHHS labels and used pseudo labels
or no stage labels, respectively, for MESA and CFS training.
Both alternatives preserve much of the performance obtained
with full labels, indicating that expert annotations are not
required for every pretraining cohort.
Even without stage guidance, WSC sleep staging performance
decreases only modestly, although larger drops occur in OSA
classification accuracy and AF recognition under LP.
The AF gaps narrow after FT.
These results suggest that SleepXA can adapt to incomplete
expert annotations through pseudo-label or partial-label guidance,
while the impact of removing stage guidance varies across
downstream tasks (Appendix~\ref{sec:stage_label_effect}).

\vspace{-10pt}
\paragraph{Robustness to modality and supervision constraints.}
Finally, we evaluate robustness to reduced sensor availability and limited downstream supervision. 
SleepXA remains effective with reduced PSG inputs, including EEG-only input. 
We further compare single-domain SHHS, matched multi-domain, and full multi-domain pre-training under different downstream labeling ratios. 
Full multi-domain pre-training most consistently benefits sleep-domain transfer, whereas gains on more distant tasks are more dependent on downstream supervision and domain shift.
Details are in Appendix~\ref{app:robustness_analysis}.

\vspace{-10pt}
\section{Conclusion}
\vspace{-10pt}
We introduced SleepXA, a compositional framework that uses synchronized
PSG to integrate independently pretrained physiological
foundation models.
SleepXA combines an asymmetric architecture with conditional common-specific
alignment to capture cross-modal common structure while preserving
modality-specific physiology. Across 27 downstream tasks spanning sleep representation, clinical prediction,
and daytime physiological transfer, SleepXA demonstrates strong transfer across
cohorts, sensor configurations, and adaptation settings.
Controlled comparisons and ablations further show that the gains arise not
only from pretrained physiological knowledge, but also from how heterogeneous
pretrained representations are composed. These results support a compositional alternative to training multimodal
physiological foundation models from scratch: synchronized PSG can provide a
common observational basis for aligning existing physiological foundation
models into transferable multisystem representations.

\subsection*{AI use statement}

In this work, we used generative AI tools to assist with language editing and the preparation of visual materials. Specifically, large language models (LLMs) were used to polish the language of initial manuscript drafts, including improving grammar, clarity, and presentation, without altering the underlying scientific claims or conclusions. We also used generative AI tools to create illustrative visual elements that were subsequently incorporated as components of the manuscript figures, such as the sleep-scene illustration shown in Figure~1.

All AI-assisted text was reviewed and revised by the authors. AI-generated visual elements were also manually inspected, selected, and integrated into the final figures by the authors to ensure that they accurately represented the intended concepts and did not introduce misleading scientific information. The experimental results, quantitative analyses, and scientific conclusions reported in this work were determined and verified by the authors.

We take responsibility for the final content of this work, including text, claims, and artifacts produced with the aid of generative AI.

\subsection*{Ethics Statement}

This study conducts secondary analysis of existing de-identified physiological and polysomnographic datasets obtained from established public or controlled-access research repositories. No new participants were recruited, no new human-subject data were collected, and no intervention or direct interaction with participants was conducted as part of this work. Ethical approval, informed consent, and participant privacy protections for the original data collection were obtained by the institutions responsible for the respective studies. We accessed and used all datasets in accordance with their licenses, data-use agreements, and repository-specific access requirements, including the National Sleep Research Resource (NSRR), PhysioNet, and other official data repositories. No attempt was made to re-identify individual participants.

\subsection*{Reproducibility Statement}

To support reproducibility, the appendix describes the dataset preprocessing,
subject-level data splits, evaluation protocols, implementation details, and
key training settings used in our experiments.

\bibliography{iclr2027_conference}
\bibliographystyle{iclr2027_conference}

\renewcommand{\thefigure}{A.\arabic{figure}}
\renewcommand{\thetable}{A.\arabic{table}}

\setcounter{figure}{0}
\setcounter{table}{0}
\newpage
\appendix
\section{Appendix}
\subsection{Datasets and Evaluation Protocols}
\label{app:datasets}
\vspace{-6pt}
We evaluate SleepXA using a broad downstream benchmark spanning
general sleep representation, clinical and disease prediction, and
daytime physiological transfer.
Table~\ref{tab:dataset_configuration} summarizes the dataset-level
configuration, including signal modalities, sample sizes, and evaluation
protocols, while Table~\ref{tab:task_definitions} provides the definitions
and prediction targets of the 27 downstream tasks.

For multimodal alignment pre-training, we use three large synchronized
polysomnography (PSG) cohorts:
the Sleep Heart Health Study (SHHS; Visits 1 and 2)
~\citep{quan1997sleep},
the Multi-Ethnic Study of Atherosclerosis (MESA)
~\citep{chen2015racial}, and
the Cleveland Family Study (CFS)
~\citep{redline1995familial}.
To prevent subject leakage across longitudinal SHHS visits, we first hold out
Test SHHS1/2, comprising 1,000 participants with PSG recordings
available at both SHHS Visits 1 and 2 (2,000 recordings in total), from all
multimodal pre-training.
After this exclusion, alignment pre-training uses 9,230 overnight PSG
recordings.
Test SHHS1/2 is reserved exclusively for downstream evaluation and provides
the held-out SHHS population used for demographic prediction, sleep analysis,
OSA assessment, and current and longitudinal clinical tasks.

We further evaluate generalization on multiple external sleep cohorts that
are never used during alignment pre-training.
These include APPLES~\citep{quan2011association},
the Wisconsin Sleep Cohort (WSC)~\citep{young2009burden},
SleepEDF20~\citep{kemp2000analysis},
HMC~\citep{alvarez2021inter},
ISRUC-1~\citep{khalighi2016isruc},
MASS-SS2~\citep{o2014montreal}, and
the CAP Sleep Database~\citep{terzano2001atlas}.
Together, these cohorts provide complementary evaluations of sleep staging,
respiratory-event detection, OSA severity estimation, sleep-spindle
detection, and sleep-disorder classification under different populations,
recording configurations, and cohort sizes.

Daytime physiological transfer is evaluated on seven additional
physiological-signal datasets, none of which are used during alignment
pre-training.
These comprise major depressive disorder (MDD) diagnosis from EEG
~\citep{mumtaz2017electroencephalogram},
dementia classification including Alzheimer's disease (AD) from resting-state EEG
using OpenNeuro ds004504~\citep{miltiadous2023dataset},
mental-arithmetic (MAT) recognition from EEG
~\citep{zyma2019electroencephalograms},
motor-imagery classification using BCI Competition IV-2a
~\citep{tangermann2012review},
atrial-fibrillation (AF) recognition from single-lead ECG using the
PhysioNet/Computing in Cardiology 2017 Challenge dataset
~\citep{clifford2017af},
constant-force hand-gesture recognition from surface EMG using
DS1~\citep{zuniga2025surfaceemg}, and
vigilance estimation using SEED-VIG
~\citep{zheng2017multimodal}.
SEED-VIG is evaluated under both EEG-only and EEG+EOG settings, yielding
eight transfer settings across the seven datasets.
We use \textit{daytime physiological transfer} as a compact umbrella term
for physiological downstream tasks outside nocturnal PSG settings,
including awake-state, ambulatory, and task-specific recordings; the term
does not imply that every recording, such as those in CinC 2017, was
explicitly acquired during daytime hours.

The downstream benchmark is organized into three groups.
\textbf{General Sleep Representation Evaluation} probes fundamental sleep
representations through demographic prediction, sleep staging,
apnea/hypopnea detection, and sleep-spindle detection, using both
Test SHHS1/2 and external sleep cohorts.
\textbf{Clinical and Disease Prediction} evaluates OSA severity,
sleep-disorder classification, current disease status, and longitudinal
prediction of subsequent sleep-related and cardiovascular outcomes.
Finally, \textbf{Daytime Physiological Transfer} evaluates cross-context transfer from overnight PSG to independent non-sleep physiological tasks across EEG, ECG, EOG, and surface-EMG settings. When participant identifiers are available, we split the data by participant, keeping all recordings and sessions from the same participant in the same split or fold. Larger cohorts use fixed 8:1:1 train/validation/test splits, while smaller cohorts are evaluated using participant-level k-fold cross-validation.

\begin{table*}[t]
\centering
\caption{
\textbf{Dataset configuration for alignment pre-training and downstream evaluation.}
Task IDs correspond to Table~\ref{tab:task_definitions}.
}
\label{tab:dataset_configuration}

\begingroup
\small
\setlength{\tabcolsep}{6.2pt}
\renewcommand{\arraystretch}{1.12}

\begin{tabular*}{\textwidth}{
@{\extracolsep{\fill}}
lccrrc@{}
}
\toprule
\textbf{Dataset}
& \textbf{Task ID(s)}
& \textbf{Signal}
& \textbf{Records/Nights}
& \textbf{Subjects}
& \textbf{Split} \\
\midrule

\multicolumn{6}{@{}l}{\textit{Alignment pre-training}} \\
\addlinespace[3pt]
SHHS-1/2
& --
& PSG
& 6,444
& 4,797
& 5,798 / 646 \\

MESA
& --
& PSG
& 2,056
& 2,056
& 1,850 / 206 \\

CFS
& --
& PSG
& 730
& 730
& 657 / 73 \\

\addlinespace[3pt]
\midrule
\multicolumn{6}{@{}l}{\textit{Downstream evaluation}} \\
\addlinespace[3pt]

Test SHHS1/2
& T1--T4, T6
& PSG
& 2,000
& 1,000
& 8:1:1 \\

APPLES
& T3--T4
& PSG
& 1,096
& 1,096
& 8:1:1 \\

WSC
& T3--T4, T6
& PSG
& 2,534 / 2,570
& 1,121 / 1,123
& 8:1:1 \\

\addlinespace[2pt]

SleepEDF20
& T3
& PSG
& 39
& 20
& 5-fold \\

ISRUC-1
& T3
& PSG
& 100
& 100
& 5-fold \\

HMC
& T3
& PSG
& 151
& 151
& 5-fold \\

MASS-SS2
& T5
& EEG
& 19
& 19
& 5-fold \\

CAP
& T7
& PSG
& 108
& 108
& 5-fold \\

\addlinespace[2pt]

Test SHHS1/2
& T8--T13
& PSG
& 2,000
& 1,000
& 5-fold \\

Test SHHS1 $\rightarrow$ SHHS2
& T14--T15
& PSG
& 1,000 pairs
& 1,000
& 5-fold \\

Test SHHS2
& T16--T19
& PSG
& 1,000
& 1,000
& 5-fold \\

\addlinespace[2pt]

MDD
& T20
& EEG
& 64
& 64
& 5-fold \\

OpenNeuro ds004504
& T21
& EEG
& 88
& 88
& 5-fold \\

MAT
& T22
& EEG
& --
& 36
& 5-fold \\

BCI IV-2a
& T23
& EEG
& --
& 9
& 5-fold \\

CinC 2017\textsuperscript{\dag}
& T24
& ECG
& 8,528
& --
& 5-fold \\

sEMG-DS1
& T25
& sEMG
& 2,000
& 20
& 5-fold \\

SEED-VIG
& T26--T27
& EEG / EEG+EOG
& --
& 21
& 5-fold \\

\bottomrule
\end{tabular*}

\vspace{4pt}

\begin{minipage}{\textwidth}
\footnotesize
\textit{Notes.}
Pre-training splits are reported as train/validation.
Downstream splits are participant-level unless otherwise noted. For WSC, sample sizes differ slightly across tasks because of task-specific label availability and preprocessing, and are reported for T3--T4 / T6, respectively.
\textsuperscript{\dag}CinC 2017 uses class-stratified record-level
five-fold cross-validation because reliable participant identifiers are unavailable.
\end{minipage}
\vspace{-15pt}
\endgroup
\end{table*}

\begin{table*}[t]
\centering
\caption{
\textbf{Overview of the 27 downstream tasks.}
Task IDs correspond to those used in the cross-task performance summary.
}
\label{tab:task_definitions}

\begingroup
\small
\setlength{\tabcolsep}{6.5pt}
\renewcommand{\arraystretch}{1.11}

\begin{tabular*}{0.98\textwidth}{
@{\extracolsep{\fill}}
c
l
c
l
@{}
}
\toprule
\textbf{ID}
& \textbf{Task}
& \textbf{Type}
& \textbf{Primary metric(s)} \\
\midrule

\multicolumn{4}{@{}l}{\textsc{General sleep representation}} \\
\addlinespace[2.5pt]

\textbf{T1}  & Age prediction                  & R      & MAE, RMSE, $R^2$ \\
\textbf{T2}  & Sex prediction                  & C2     & Acc., MF1, AUROC \\
\textbf{T3}  & Sleep staging                   & C5     & Acc., MF1, $\kappa$ \\
\textbf{T4}  & Apnea/hypopnea detection        & C2     & Acc., AUROC, AUPRC \\
\textbf{T5}  & Sleep-spindle detection         & D      & Precision, Recall, F1 \\

\addlinespace[3pt]
\midrule
\multicolumn{4}{@{}l}{\textsc{Clinical and disease prediction}} \\
\addlinespace[2.5pt]

\textbf{T6}  & AHI estimation / OSA severity   & R+C4   & MAE, RMSE, $R^2$; Acc., MF1 \\
\textbf{T7}  & Sleep-disorder classification   & C2     & MF1, AUROC \\

\addlinespace[1.5pt]

\textbf{T8}  & COPD                            & C2     & AUROC \\
\textbf{T9}  & Emphysema                       & C2     & AUROC \\
\textbf{T10} & Clinical depression             & C2     & AUROC \\
\textbf{T11} & Depressive affect               & C2     & AUROC \\
\textbf{T12} & Heart failure                   & C2     & AUROC \\
\textbf{T13} & Stroke                          & C2     & AUROC \\

\addlinespace[1.5pt]

\textbf{T14} & AHI worsening                   & C2     & AUROC \\
\textbf{T15} & Incident hypersomnia            & C2     & AUROC \\
\textbf{T16} & Composite CVD                   & C2     & AUROC \\
\textbf{T17} & Incident heart failure          & C2     & AUROC \\
\textbf{T18} & Incident stroke                 & C2     & AUROC \\
\textbf{T19} & CVD death                       & C2     & AUROC \\

\addlinespace[3pt]
\midrule
\multicolumn{4}{@{}l}{\textsc{Daytime physiological transfer}} \\
\addlinespace[2.5pt]

\textbf{T20} & MDD diagnosis                   & C2     & AUROC \\
\textbf{T21} & Dementia classification         & C3     & MF1 \\
\textbf{T22} & Mental-arithmetic recognition   & C2     & MF1 \\
\textbf{T23} & Motor-imagery classification    & C4     & MF1 \\
\textbf{T24} & Atrial-fibrillation recognition & C4     & Ch-F1, AF-F1, AUROC \\
\textbf{T25} & Hand-gesture recognition        & C5     & MF1 \\
\textbf{T26} & Vigilance estimation (EEG)      & R      & Pearson $r$ \\
\textbf{T27} & Multimodal vigilance estimation (EEG+EOG)  & R      & Pearson $r$ \\

\bottomrule
\end{tabular*}

\vspace{4pt}
\begin{minipage}{0.98\textwidth}
\footnotesize
\textit{Notes.}
R denotes regression, C$k$ denotes $k$-class classification, and D denotes
event detection. T6 jointly evaluates continuous AHI estimation and
four-level OSA severity classification. T24 is a four-class ECG rhythm classification task (Normal, AF, Other, and Noisy). Ch-F1 is the official CinC 2017 score averaged over Normal, AF, and Other; AF-F1 and AUROC evaluate AF recognition. T26 and T27 share the same vigilance target but differ in input modality:
T26 uses EEG only, whereas T27 uses EEG+EOG to assess
multimodal fusion capability in downstream transfer specifically.
\end{minipage}
\vspace{-15pt}
\endgroup
\end{table*}

\begin{table*}[t]
\centering
\caption{
\textbf{Pre-training provenance of inherited EEG and ECG foundation encoders.}
LaBraM and CSFM have no cohort overlap with our downstream evaluation datasets.
}
\vspace{3pt}
\label{tab:fm_pretraining_overlap}

\begingroup
\small
\setlength{\tabcolsep}{5pt}
\renewcommand{\arraystretch}{1.15}

\begin{tabularx}{\textwidth}{
@{}
l
l
>{\raggedright\arraybackslash}X
>{\centering\arraybackslash}p{2.3cm}
>{\centering\arraybackslash}p{1.8cm}
@{}
}
\toprule
\textbf{Model}
&
\textbf{Signal}
&
\textbf{Original Pre-training Corpus}
&
\textbf{Pre-training Scale}
&
\textbf{Overlap}
\\
\midrule

LaBraM
&
EEG
&
20 heterogeneous EEG datasets spanning BCI, emotion,
resting-state, and clinical EEG, including the SEED and TUH series,
together with other public and self-collected EEG datasets.
&
2,534.78 h
&
\textbf{None}
\\

\addlinespace[4pt]

CSFM
&
ECG/PPG
&
MIMIC-III-WDB, MIMIC-IV-ECG, and CODE-Full.
&
$\sim$1.7M individuals
&
\textbf{None}
\\

\bottomrule
\end{tabularx}

\vspace{3pt}
\begin{minipage}{\textwidth}
\footnotesize
\textit{Note.}
``Overlap'' denotes direct reuse of the same source cohort or recordings
between foundation-model pre-training and our downstream evaluation.
\end{minipage}
\vspace{-20pt} 
\endgroup
\end{table*}

\vspace{-4pt}
\subsection{Implementation Details}
\label{app:implementation}
\paragraph{Pre-training modalities and signal preprocessing.}
SleepXA performs multimodal alignment over seven physiological modalities
commonly recorded in overnight polysomnography (PSG):
electroencephalography (EEG), electrooculography (EOG),
electromyography (EMG), electrocardiography (ECG), airflow,
abdominal respiratory effort (ABD), and pulse oximetry (SpO$_2$).
All PSG signals are temporally synchronized and segmented into non-overlapping
30-s epochs. EEG, EOG, EMG, and ECG are first standardized to 100~Hz, airflow and ABD to 10~Hz, and SpO$_2$ to 1~Hz; EEG and ECG are subsequently resampled to the native sampling rates required by their pretrained backbones.
Each signal is normalized independently by channel, with extreme normalized
values clipped to reduce the influence of outliers.
Because signal availability varies across cohorts and recordings, unavailable
modalities are excluded from the corresponding alignment terms using explicit
availability masks rather than being imputed from other signals.
\vspace{-4pt}
\paragraph{Modality encoders.}
The EEG and ECG branches are initialized from pretrained LaBraM and CSFM
encoders, respectively (Table~\ref{tab:fm_pretraining_overlap}).
The remaining modalities, including EOG, EMG, airflow, ABD, and SpO$_2$, are processed by independently parameterized lightweight 1D CNN encoders, respectively.
Each lightweight encoder consists of three convolutional layers,
$\mathrm{Conv1D}(1,32,k=25,s=5)$,
$\mathrm{Conv1D}(32,64,k=15,s=3)$, and
$\mathrm{Conv1D}(64,128,k=9,s=2)$,
followed by global average pooling to produce a 128-dimensional
representation.
Temporal padding is applied where required for low-rate inputs.

\vspace{-4pt}
\paragraph{Optimization and hyperparameters.}
All optimization settings are summarized here for reproducibility.
Multimodal alignment pre-training is performed for up to 50 epochs using
AdamW with a learning rate of $3\times10^{-5}$, weight decay
$1\times10^{-4}$, linear learning-rate warm-up, and cosine decay.
Training uses BF16 mixed precision on four NVIDIA H200 GPUs with a batch size
of 512 per GPU, corresponding to a global batch size of 2,048.
During alignment pre-training, the lower layers of the pretrained LaBraM and
CSFM encoders remain frozen, while their upper layers are progressively
unfrozen.
The lightweight modality encoders and alignment modules are optimized
throughout pre-training.
Validation is performed periodically according to optimization steps, and the
checkpoint with the lowest validation alignment loss is retained.
Early stopping uses a patience of 10 validation evaluations.
For downstream adaptation, LP uses a learning rate of $1\times10^{-4}$ for the
prediction head.
FT uses learning rates of $1\times10^{-4}$ and $1\times10^{-6}$ for the
prediction head and backbone, respectively, corresponding to a backbone
learning-rate scale of 0.01.
Both settings use AdamW with weight decay $1\times10^{-4}$.
Downstream models are generally trained for up to 50 epochs
with a batch size of 32 and an early-stopping patience of 4.
Downstream experiments are also conducted using NVIDIA H200 GPUs. We typically use a fixed random seed of 42 for both alignment pre-training and downstream evaluation, with repeated-seed experiments explicitly indicated where applicable.

SleepXA's total parameter count depends on the EEG and ECG foundation models used. Our main configuration combines LaBraM and CSFM, with SleepXA's multimodal alignment modules adding 6.81M parameters (5.57\% of the total). The table~\ref{tab:model_size_memory} below summarizes model size and measured GPU memory usage during inference. For inference, we evaluated the main SleepXA configuration on the WSC PSG recordings using a single NVIDIA H200 GPU. SleepXA achieved a throughput of 1.93 recordings per second. Its measured peak reserved GPU memory was 0.95 GiB, approximately 44.4\% lower than SleepGPT and 38.7\% lower than OSF in the evaluated configurations.

\begin{table}[htbp]
\centering
\small
\caption{\textbf{Model size and inference GPU memory usage.}}
\label{tab:model_size_memory}
\setlength{\tabcolsep}{6pt}
\renewcommand{\arraystretch}{1.10}
\begin{tabular}{@{}lcc@{}}
\toprule
\textbf{Model}
& \shortstack{\textbf{Total parameters}\\\textbf{(M)}}
& \shortstack{\textbf{Peak reserved GPU}\\\textbf{memory (GiB)}} \\
\midrule
SleepXA (LaBraM + CSFM) & 122.25 & \textbf{0.95} \\
OSF                         & 85.33  & 1.55 \\
SleepGPT                    & 134.80 & 1.71 \\
\bottomrule
\end{tabular}
\end{table}

\subsection{Baselines and Downstream Evaluation Protocol}
\label{app:baselines}

\paragraph{Baseline methods}
We organize the compared methods into three groups.

(1) \textbf{Non-foundation models} include task-specific supervised
architectures, general-purpose time-series models, and conventional
self-supervised representation-learning methods.
We select baselines according to the characteristics of each task, prioritizing strong domain-specific models with public implementations when available and using a unified set of general-purpose baselines otherwise.

The general-purpose baselines are selected following representative benchmark
suites used in LaBraM~\citep{jiang2024large},
BrainOmni~\citep{xiao2026brainomni}, BIOT~\citep{ yang2023biot} and
CSFM~\citep{gu2026cardiac}.
Depending on the input modality and task, these include
ST-Transformer architectures for EEG~\citep{song2021transformer},
CNN--Transformer models~\citep{peh2022transformer},
ResNet1D-based architectures commonly used for ECG analysis~\citep{ribeiro2020automatic,gu2026cardiac},
general time-series models including
TimesNet~\citep{wu2022timesnet} and
iTransformer~\citep{liu2024itransformer}, and
ContraWR~\citep{yang2023self} as a conventional self-supervised EEG
representation-learning baseline.

For sleep-related tasks, we additionally include representative
domain-specific supervised methods.
For sleep staging, we compare against
DeepSleepNet~\citep{supratak2017deepsleepnet},
SleePyCo~\citep{lee2024sleepyco},
U-Time~\citep{perslev2019u},
AttnSleep~\citep{eldele2021attention},
XSleepNet~\citep{phan2021xsleepnet},
ASTGSleep~\citep{chen2025astgsleep},
XSleepFusion~\citep{hu2025xsleepfusion}, and
SleepXViT~\citep{lee2025explainable}.
For respiratory event detection, we include
Hybrid Transformer~\citep{hu2022hybrid},
XSleepFusion~\citep{hu2025xsleepfusion}, and
IPCT-Net~\citep{hu2025ipct}.
For OSA severity estimation, because publicly available OSA-specific baselines are limited, we use the unified general baseline set above and additionally include TSDNet~\citep{hu2025transparent} as an OSA-specific supervised method.
For sleep spindle detection on MASS-SS2, we compare against
RED-Time~\citep{tapia2020red} and
SUMO~\citep{kaulen2022advanced}.

(2) \textbf{Physiological foundation models} include
LaBraM~\citep{jiang2024large} and
CBraMod~\citep{wang2025cbramod} for EEG-based evaluation, and
CSFM~\citep{gu2026cardiac} for ECG-based evaluation.
These baselines assess whether the proposed multimodal alignment provides
additional benefits beyond large-scale unimodal physiological pretraining.

(3) \textbf{Sleep foundation models} include
SleepFM~\citep{thapa2026sleepfm},
SleepGPT~\citep{huang2026sleepgpt}, and
OSF~\citep{shuai2026osf}.
We select these three major sleep foundation models as representative
baselines because they
cover recent and complementary approaches to sleep representation learning
while providing publicly available pretrained weights and sufficiently
documented downstream evaluation pipelines for reproducible comparison.
Our goal is not to exhaustively enumerate all existing sleep foundation
models, but to establish a representative and reproducible comparison set
spanning different pretraining paradigms and input configurations.
Whenever applicable, we use their officially released checkpoints, supported
input modalities, preprocessing pipelines, and downstream adaptation
protocols.

\paragraph{Common downstream evaluation protocol.}

We evaluate pretrained foundation models using linear probing (LP) and
full fine-tuning (FT). In LP, the pretrained backbone is frozen, and only a
task-specific linear prediction head is trained. In FT, the prediction
head and the backbone layers selected by the fine-tuning schedule are
optimized jointly. Epoch-level tasks use individual epoch
representations, whereas whole-night and subject-level tasks aggregate
epoch representations according to their task protocols.

\paragraph{Cross-task ranking protocol.} To emphasize pretrained representation quality, foundation models are ranked primarily by LP performance, which evaluates downstream utility without modifying the pretrained backbone; FT is used only to break LP ties. A supervised baseline is ranked ahead of a foundation model only when it surpasses both its LP and FT performance, so that task-specific supervised training is considered superior only when it also exceeds the fully adapted foundation model. For each task, ranking is based on its primary evaluation metric, such as Macro-F1 for classification tasks, AUROC for disease prediction tasks, and the corresponding regression metric (e.g., MAE or Pearson’s \(r\)) for continuous prediction tasks.

For models with public implementations or pretrained checkpoints, we
follow their documented configurations and adapt inputs as required.
Downstream signals undergo dataset-specific preprocessing, including channel selection, filtering, resampling, normalization, and sample construction where appropriate, following established evaluation protocols from prior foundation-model studies~\citep{jiang2024large,xiao2026brainomni}, and the preprocessing recommendations provided with each dataset. Within each task, compared methods use the same splits, prediction targets, and evaluation metrics. Input modalities follow each baseline's supported configuration; for the sEMG task, EEG foundation models are used as the closest available baselines due to the lack of comparable pretrained sEMG foundation models.

\paragraph{Evaluation metrics.}
We use task-specific evaluation metrics following commonly adopted protocols
for the corresponding benchmarks and prior literature.
For sleep staging, each 30-s epoch is classified into five AASM stages:
Wake, N1, N2, N3, and REM, following the AASM scoring manual
~\citep{berry2017aasm}.
We report accuracy (Acc), macro-F1 (MF1), and Cohen's $\kappa$, which measure
overall accuracy, class-balanced performance, and agreement beyond chance,
respectively
~\citep{supratak2017deepsleepnet,phan2021xsleepnet,eldele2021attention}. For respiratory-event detection, we report area under the precision--recall
curve (AUPRC) and area under the receiver operating characteristic curve
(AUROC), which are informative under class imbalance, together with accuracy
for comparison with prior sleep-event detection studies. For obstructive sleep apnea (OSA) assessment, the apnea--hypopnea index (AHI)
is evaluated both as a continuous clinical measure and through conventional
severity categories defined by AHI thresholds~\citep{berry2017aasm}.
Accordingly, we report mean absolute error (MAE), root mean squared error
(RMSE), and $R^2$ for AHI regression, together with accuracy and macro-F1 for
OSA severity classification
~\citep{hu2025transparent}.

For other disease and clinical classification tasks, we primarily use AUROC for imbalanced binary outcomes and MF1 for multiclass classification benchmarks. Regression and continuous physiological prediction tasks use the corresponding
standard regression or correlation metrics.
Metrics for all 27 downstream tasks are also summarized in
Table~\ref{tab:task_definitions}.

\vspace{-4pt}
\subsection{Additional Results and Analyses}
\label{aap：Additional Results and Analyses}
\vspace{-4pt}
\subsubsection{Detailed Results for General Sleep Representation}
\label{app:detailed_downstream_results}
We provide additional analyses for the tasks
summarized in the main text.
Results already reported in full in the main text are omitted to avoid repetition.
\vspace{-5pt}
\paragraph{Age and Sex Prediction.}
Table~\ref{tab:age_sex_results} reports the complete results for subject-level
age and sex prediction on Test SHHS1/2.
For age estimation, SleepXA with full PSG achieves the best performance under
both LP and FT. Its LP results (MAE 5.250, RMSE 6.680, $R^2=0.584$) are
comparable to, and slightly better than, the corresponding FT results,
indicating that age-related information is already well represented before
downstream adaptation.
For sex classification, OSF performs best under LP, while SleepXA with full
PSG achieves the strongest FT accuracy and MF1 and remains competitive
in AUROC. Several baselines show modest degradation after FT, suggesting that end-to-end adaptation can be sensitive to model architecture and task-specific optimization, whereas SleepXA remains comparatively stable across LP and FT.

\begin{table}[!ht]
\centering
\caption{
\textbf{Age prediction and sex classification on Test SHHS1/2.}
Lower values are better for MAE and RMSE, while higher values are better
for $R^2$, Acc, MF1, and AUROC.
\textbf{Bold} and \underline{underlined} values indicate the
\textbf{best} and \underline{second-best} model results.
}
\label{tab:age_sex_results}

\begingroup
\footnotesize
\setlength{\tabcolsep}{2.5pt}
\renewcommand{\arraystretch}{0.92}

\begin{tabularx}{0.97\linewidth}{
@{}lc*{6}{C}@{}
}
\toprule[1.0pt]

\multirow{2}{*}{\textbf{Model}}
& \multirow{2}{*}{\textbf{Channels}}
& \multicolumn{3}{c}{\textbf{Age prediction}}
& \multicolumn{3}{c}{\textbf{Sex classification}} \\

\cmidrule(lr){3-5}
\cmidrule(lr){6-8}

&
& \textbf{MAE} $\downarrow$
& \textbf{RMSE} $\downarrow$
& $\mathbf{R^2}$ $\uparrow$
& \textbf{Acc.} $\uparrow$
& \textbf{MF1} $\uparrow$
& \textbf{AUROC} $\uparrow$ \\

\midrule
\rowcolor{SectionGray}
\multicolumn{8}{@{}l}{
    \textit{General non-foundation models}
} \\
\addlinespace[0.5pt]

TimesNet{\scriptsize~\citep{wu2022timesnet}} & Full
& 7.565 & 9.328 & 0.268
& 0.725 & 0.724 & 0.780 \\

CNN-Transformer{\scriptsize~\citep{peh2022transformer}} & Full
& 7.063 & 8.964 & 0.324
& 0.490 & 0.329 & 0.494 \\

ResNet1D-50{\scriptsize~\citep{ribeiro2020automatic}}
 & Full
& 7.375 & 9.642 & 0.218
& 0.585 & 0.584 & 0.579 \\

ST-Transformer{\scriptsize~\citep{song2021transformer}}
 & Full
& 7.005 & 9.163 & 0.294
& 0.610 & 0.600 & 0.690 \\

ContraWR{\scriptsize~\citep{yang2023self}} & Full
&6.731 & 8.541 & 0.386
& 0.675 & 0.674
& 0.718 \\

iTransformer{\scriptsize~\citep{liu2024itransformer}} & Full
& 8.292 & 10.312 & 0.106
& 0.570 & 0.570 & 0.606 \\

\midrule
\rowcolor{LPBlue}
\multicolumn{8}{@{}l}{
    \textit{Foundation models: linear probing (LP)}
} \\
\addlinespace[1pt]

LaBraM{\scriptsize~\citep{jiang2024large}}
 & E
& 9.917 & 11.781 & -0.292
& 0.590 & 0.590 & 0.628 \\

CBraMod{\scriptsize~\citep{wang2025cbramod}} & E
& 8.813 & 10.827 & 0.014
& 0.540 & 0.539 & 0.577 \\

CSFM{\scriptsize~\citep{gu2026cardiac}} & C
& 7.777 & 9.735 & 0.118
& 0.561 & 0.544 & 0.635 \\

SleepFM{\scriptsize~\citep{thapa2026sleepfm}}
 & Full
& 8.227 & 9.906 & 0.086
& 0.717 & 0.717 & 0.790 \\

SleepGPT{\scriptsize~\citep{huang2026sleepgpt}} & Full
& 6.146 & 7.765 & 0.410
& 0.636 & 0.631 & 0.716 \\

OSF{\scriptsize~\citep{shuai2026osf}} & Full
& 6.124 & 7.561 & 0.468
& 0.813 & 0.812 & 0.889 \\

\rowcolor{LPBlue}
\textbf{SleepXA (Ours)} & Full
& \textbf{5.250} & \textbf{6.680} & \textbf{0.584}
& 0.793 & 0.792 & 0.876 \\

\midrule
\rowcolor{FTPurple}
\multicolumn{8}{@{}l}{
    \textit{Foundation models: full fine-tuning (FT)}
} \\
\addlinespace[1pt]

LaBraM{\scriptsize~\citep{jiang2024large}} & E
& 8.059 & 9.763 & 0.112
& 0.699 & 0.693 & 0.760 \\

CBraMod{\scriptsize~\citep{wang2025cbramod}}
 & E
& 6.168 & 8.018 & 0.459
& 0.645 & 0.645 & 0.723 \\

CSFM{\scriptsize~\citep{gu2026cardiac}} & C
& 7.374 & 9.330 & 0.189
& 0.772 & 0.771 & 0.857 \\

SleepFM{\scriptsize~\citep{thapa2026sleepfm}} & Full
& 7.442 & 9.098 & 0.229
& 0.711 & 0.711 & 0.790 \\

SleepGPT{\scriptsize~\citep{huang2026sleepgpt}} & Full
& 8.047 & 9.760 & 0.113
& 0.762 & 0.762 & 0.838 \\

OSF{\scriptsize~\citep{shuai2026osf}} & Full
& 7.075 & 8.801 & 0.279
& \underline{0.839} & \underline{0.838} & \underline{0.912}\\

\rowcolor{FTPurple}
\textbf{SleepXA (Ours)} & Full
& \underline{5.277} & \underline{6.697} & \underline{0.582}
& \textbf{0.879} & \textbf{0.878} & \textbf{0.939} \\

\bottomrule[1.0pt]
\end{tabularx}

\par\vspace{2pt}
\begin{minipage}{\linewidth}
\scriptsize
    E: EEG; C: ECG; E/O/M: EEG+EOG+EMG; Full: full experimental PSG configuration.
\end{minipage}

\endgroup
\end{table}

\begin{figure*}[t]
    \centering
    \includegraphics[
        width=\textwidth
    ]{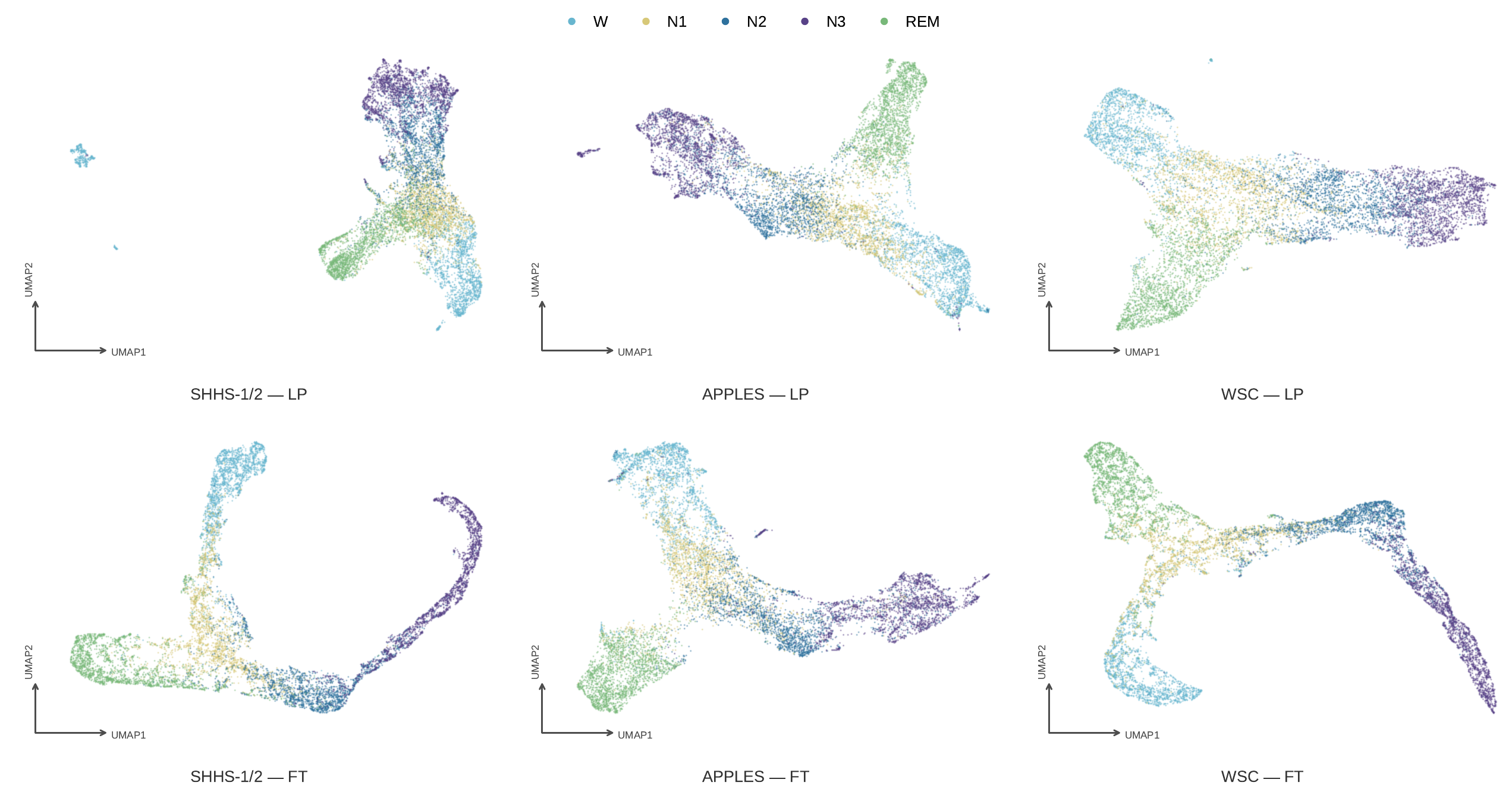}
    \vspace{-6pt}
    \caption{
    \textbf{UMAP visualization of sleep-stage representations.}
    Epoch-level representations from Test SHHS1/2, APPLES, and WSC are shown
    under LP (top) and FT (bottom).
    Points are colored according to the five sleep stages:
    Wake, N1, N2, N3, and REM.
    }
    \label{fig:sleep_stage_umap}
    \vspace{-10pt}
\end{figure*}

\vspace{-10pt}
\paragraph{Sleep Stage Representation Visualization.}
\label{app:sleep_representation}
To complement the quantitative sleep-staging results in the main text,
we qualitatively examine the organization of the learned representations
across Test SHHS1/2, APPLES, and WSC.
Figure~\ref{fig:sleep_stage_umap} visualizes epoch-level embeddings using
UMAP under both LP and FT.
Across the three cohorts, the representations exhibit consistent
stage-dependent organization while retaining overlap between neighboring
sleep states, particularly around transitional stages such as N1.
After full fine-tuning, the embedding geometry becomes more structured with
respect to the downstream staging labels.
These visualizations provide a qualitative complement to the quantitative
cross-cohort results and suggest that the learned sleep-stage structure is
preserved across independent cohorts.

\vspace{-5pt}
\paragraph{Sleep Staging on Small Cohorts.}
Beyond the large-cohort evaluations in the main text, we further assess
transfer to three widely used small-cohort sleep-staging benchmarks:
SleepEDF20~\citep{kemp2000analysis},
HMC~\citep{alvarez2021inter}, and ISRUC-1~\citep{khalighi2016isruc}.
All experiments use subject-level five-fold cross-validation to evaluate
generalization to unseen participants in these small cohorts. For supervised baselines, we include DeepSleepNet as a classical
single-channel sleep-staging model~\citep{supratak2017deepsleepnet}, XSleepNet ~\citep{phan2021xsleepnet}as a representative
multichannel sleep-staging architecture, and ASTGSleep as a full-channel
graph-based fusion model~\citep{chen2025astgsleep}, providing complementary comparisons with the
evaluated sleep foundation models. As shown in Table~\ref{tab:sleep_staging_results}, SleepXA remains competitive
under LP and achieves strong performance after FT across all
three cohorts.
In particular, FT reaches
86.76\% accuracy, 81.50\% Macro-F1, and $\kappa=0.816$ on SleepEDF20;
77.44\%, 76.28\%, and $0.708$ on HMC; and
79.07\%, 77.10\%, and $0.732$ on ISRUC-1, respectively.
These results extend the large-cohort findings in the main text and show that
the learned representation remains transferable under smaller downstream
cohorts, subject-level cross-validation, and different sensor configurations.

\begin{figure*}[t]
    \centering
    \includegraphics[
        width=\textwidth
    ]{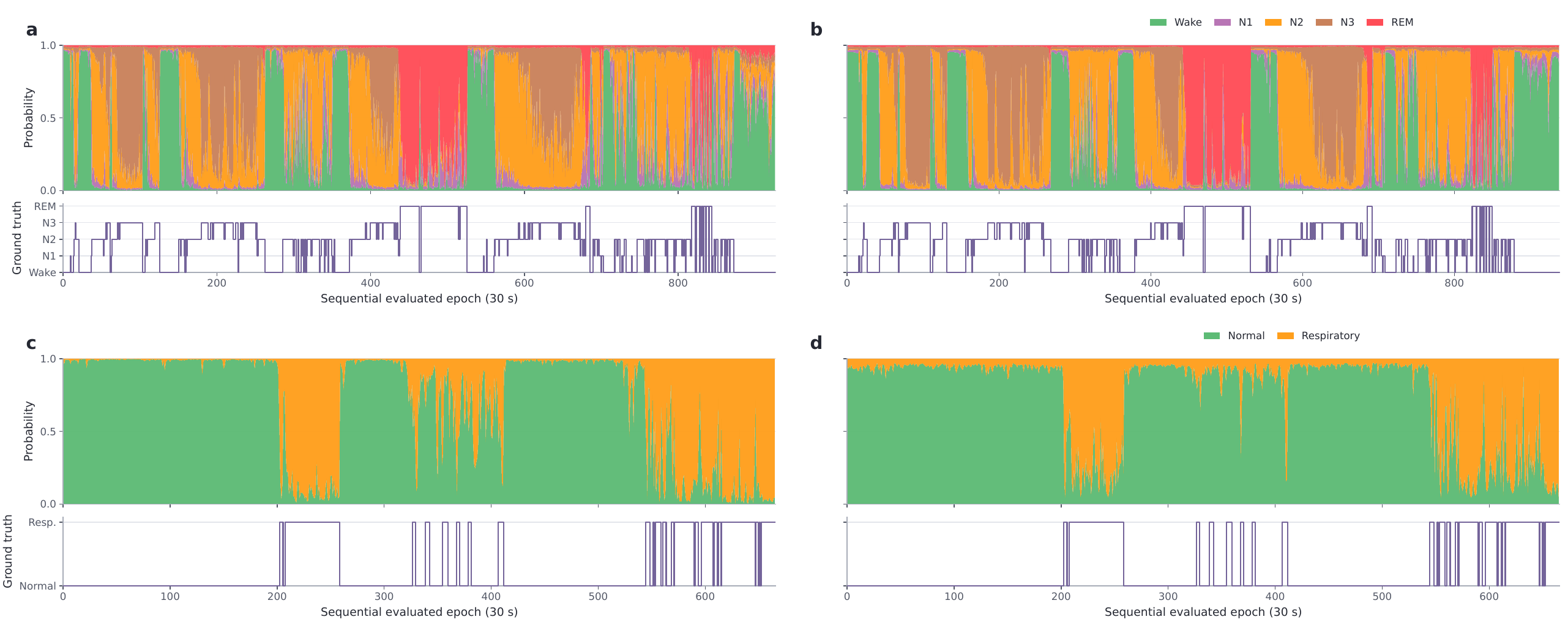}
    \vspace{-6pt}
   \caption{
\textbf{Illustrative whole-night predictions on a randomly selected WSC test recording.}
Recording ID: \texttt{39201}.
Panels show sleep staging under (a) LP and (b) FT, and respiratory-event
detection under (c) LP and (d) FT.
Prediction probabilities are shown with the corresponding ground-truth labels.
Unevaluated epochs are removed and the remaining epochs are sequentially
reindexed.
}
\label{fig:whole_night_prediction}
    \vspace{-10pt}
\end{figure*}

Fig.~\ref {fig:whole_night_prediction}(a,b) provides a complementary whole-night view of model behavior. We randomly select one recording from the WSC test set and visualize predictions across the whole night for both sleep staging and respiratory-event detection under LP and FT. For sleep staging, LP already captures the overall whole-night sleep architecture well, while FT produces more confident and temporally coherent predictions in some fragmented or transition-heavy periods.

\begin{table*}[t]
\centering
\caption{
\textbf{Sleep staging on small cohorts under subject-level five-fold
cross-validation.}
Acc. and MF1 are percentages; all metrics are higher-is-better.
\textbf{Bold} and \underline{underlined} values indicate the
\textbf{best} and \underline{second-best} results.
}
\label{tab:sleep_staging_results}

\vspace{1mm}

\begingroup
\footnotesize
\setlength{\tabcolsep}{1.7pt}
\renewcommand{\arraystretch}{0.92}

\begin{tabularx}{0.97\linewidth}{
@{}lc*{9}{C}@{}
}
\toprule[1.0pt]

\multirow{2}{*}{\textbf{Model}}
& \multirow{2}{*}{\textbf{Channels}}
& \multicolumn{3}{c}{\textbf{Sleep-EDF-20}}
& \multicolumn{3}{c}{\textbf{HMC}}
& \multicolumn{3}{c}{\textbf{ISRUC-1}} \\

\cmidrule(lr){3-5}
\cmidrule(lr){6-8}
\cmidrule(lr){9-11}

&
& \textbf{Acc.}
& \textbf{MF1}
& $\boldsymbol{\kappa}$
& \textbf{Acc.}
& \textbf{MF1}
& $\boldsymbol{\kappa}$
& \textbf{Acc.}
& \textbf{MF1}
& $\boldsymbol{\kappa}$ \\

\midrule
\rowcolor{SectionGray}
\multicolumn{11}{@{}l}{
\textit{Task-specific supervised models}
} \\
DeepSleepNet{\scriptsize~\citep{supratak2017deepsleepnet}}
& E
& 81.90 & 76.60 & 0.760
& 63.06 & 59.32 & 0.521
& 69.00 & 65.22 & 0.604 \\

XSleepNet{\scriptsize~\citep{phan2021xsleepnet}}
& E/O/M
& 86.30 & 80.60 & 0.813
& 75.76 & 73.41 & 0.684
& \textbf{80.07} & \textbf{78.74} & \textbf{0.743} \\

ASTGSleep{\scriptsize~\citep{chen2025astgsleep}}
& Full
& \underline{86.50} & \underline{80.90} & \underline{0.814}
& 74.88 & 72.66 & 0.673
& 78.50 & 77.40 & 0.724 \\

\midrule

\rowcolor{LPBlue}
\multicolumn{11}{@{}l}{
\textit{Foundation models: linear probing (LP)}
} \\

SleepFM{\scriptsize~\citep{thapa2026sleepfm}}

& Full
& 76.27 & 69.89 & 0.683
& 66.53 & 64.82 & 0.570
& 72.21 & 70.98 & 0.647 \\

SleepGPT{\scriptsize~\citep{huang2026sleepgpt}}
& Full
& 81.62 & 76.48 & 0.753
& 74.18 & 72.25 & 0.665
& 71.44 & 68.62 & 0.634 \\

OSF{\scriptsize~\citep{shuai2026osf}}
& Full
& 82.18 & 77.23 & 0.760
& 75.56 & 74.50 & 0.686
& 75.62 & 73.25 & 0.687 \\
\rowcolor{LPBlue}
\textbf{SleepXA (Ours)}
& Full
& 82.32 & 77.09 & 0.761
& 76.69 & 74.83 & 0.698
& 78.63 & \underline{77.76} & 0.727 \\

\midrule

\rowcolor{FTPurple}
\multicolumn{11}{@{}l}{
\textit{Foundation models: full fine-tuning (FT)}
} \\

SleepFM{\scriptsize~\citep{thapa2026sleepfm}}
& Full
& 79.93 & 73.39 & 0.728
& 72.44 & 69.73 & 0.638
& 74.18 & 72.26 & 0.670 \\

SleepGPT{\scriptsize~\citep{huang2026sleepgpt}}
& Full
& 83.57 & 78.40 & 0.777
& \underline{77.17} & \underline{75.90} & \underline{0.704}
& 74.23 & 73.11 & 0.670 \\

OSF{\scriptsize~\citep{shuai2026osf}}
& Full
& 83.79 & 79.27 & 0.781
& 76.53 & 75.21 & 0.695
& 78.26 & 77.02 & 0.722 \\
\rowcolor{FTPurple}
\textbf{SleepXA (Ours)}
& Full
& \textbf{86.76} & \textbf{81.50} & \textbf{0.816}
& \textbf{77.44} & \textbf{76.28} & \textbf{0.708}
&   \underline{79.07} & 77.10 & \underline{0.732} \\

\bottomrule[1.0pt]
\end{tabularx}

\vspace{1mm}

\begin{minipage}{0.99\textwidth}
\scriptsize
E: EEG; E/O/M: EEG+EOG+EMG; Full: full experimental PSG configuration.
\end{minipage}
\vspace{-10pt} 
\endgroup

\vspace{-1mm}
\end{table*}

\paragraph{Respiratory Event Detection.}
We further evaluate whether the learned representations capture
epoch-level respiratory abnormalities across cohorts.
Unlike subject-level AHI estimation and OSA severity classification,
this task probes temporally localized respiratory events within the overnight
recording.
Table~\ref{tab:resp_event_results} reports results on Test SHHS1/2,
APPLES, and WSC.

Under LP, SleepXA achieves the strongest overall performance among
the evaluated foundation models across all three cohorts.
It reaches 83.44\% accuracy, 0.763 AUPRC, and 0.879 AUROC on Test SHHS1/2;
79.61\%, 0.793, and 0.881 on APPLES; and
90.73\%, 0.790, and 0.938 on WSC.
These results indicate that respiratory-event information is accessible from
the frozen representation and transfers across independent sleep cohorts. FT further improves performance on Test SHHS1/2 and APPLES and
maintains strong performance on WSC, showing that the pretrained
representation also supports task-specific adaptation. SleepMaMi also performs competitively after FT, achieving slightly higher AUPRC than SleepXA on APPLES and WSC, with comparable AUROC on both cohorts. For consistency with the sleep-staging visualization, Fig.~\ref{fig:whole_night_prediction}(c,d) shows whole-night respiratory-event probabilities under LP and FT for the same randomly selected WSC test recording.

\begin{table}[!htbp]
\centering
\caption{
\textbf{Respiratory event detection performance on three representative cohorts.}
Acc. is reported in percentage, while AUPRC and AUROC are reported on a
0--1 scale. Higher is better for all metrics.
}
\label{tab:resp_event_results}

\begingroup
\footnotesize
\setlength{\tabcolsep}{1.7pt}
\renewcommand{\arraystretch}{0.92}

\begin{tabularx}{0.97\linewidth}{
@{}lc*{9}{C}@{}
}
\toprule[1.0pt]

\multirow{2}{*}{\textbf{Model}}
& \multirow{2}{*}{\textbf{Channels}}
& \multicolumn{3}{c}{\textbf{Test SHHS1/2}}
& \multicolumn{3}{c}{\textbf{APPLES}}
& \multicolumn{3}{c}{\textbf{WSC}} \\

\cmidrule(lr){3-5}
\cmidrule(lr){6-8}
\cmidrule(lr){9-11}

&
& \textbf{Acc.}
& \textbf{AUP}
& \textbf{AUR}
& \textbf{Acc.}
& \textbf{AUP}
& \textbf{AUR}
& \textbf{Acc.}
& \textbf{AUP}
& \textbf{AUR} \\

\midrule
\rowcolor{SectionGray}
\multicolumn{11}{@{}l}{
\textit{Task-specific supervised models}
} \\
\addlinespace[0.5pt]
Hybrid Transformer{\scriptsize~\citep{hu2022hybrid}} & C
& 58.52 & 0.329 & 0.601
& 56.89 & 0.368 & 0.542
& 78.05 & 0.262 & 0.634 \\

XSleepFusion{\scriptsize~\citep{hu2025xsleepfusion}} & Full
& 81.13 & 0.758 & 0.878
& \underline{81.02} & 0.777 & 0.871
& \textbf{92.26} & \textbf{0.845} & \textbf{0.955} \\

IPCT-Net{\scriptsize~\citep{hu2025ipct}} & Full
&\underline{83.54} & \underline{0.763} & \underline{0.881}
& 80.58 & 0.771 & 0.864
& 90.77 & 0.834 & 0.931 \\

\midrule
\rowcolor{LPBlue}
\multicolumn{11}{@{}l}{
\textit{Foundation models: linear probing (LP)}
} \\
\addlinespace[2pt]

CSFM{\scriptsize~\citep{gu2026cardiac}} & C
& 67.89 & 0.340 & 0.610
& 47.21 & 0.382 & 0.551
& 74.83 & 0.246 & 0.624 \\

SleepFM{\scriptsize~\citep{thapa2026sleepfm}}
 & Full
& 77.35 & 0.608 & 0.795
& 77.44 & 0.754 & 0.852
& 89.74 & 0.740 & 0.924 \\

SleepGPT{\scriptsize~\citep{huang2026sleepgpt}}
 & Full
& 72.70 & 0.659 & 0.820
& 75.77 & 0.747 & 0.847
& 89.98 & 0.755 & 0.922 \\

SleepMaMi{\scriptsize~\citep{park2026sleepmami}} & Full
& 72.29 & 0.641 & 0.818
& 78.42 & 0.771 & 0.866
& 87.06 & 0.763 & 0.900 \\

OSF{\scriptsize~\citep{shuai2026osf}} & Full
& 83.07 & 0.760 & 0.876
& 78.78 & 0.775 & 0.868
& 89.28 & 0.733 & 0.915 \\

\rowcolor{LPBlue}
SleepXA (Ours) & Full
& 83.44 & \underline{0.763} & 0.879
& 79.61 & \underline{0.793} & \underline{0.881}
& 90.73 & 0.790 & \underline{0.938} \\

\midrule
\rowcolor{FTPurple}
\multicolumn{11}{@{}l}{
\textit{Foundation models: full fine-tuning (FT)}
} \\
\addlinespace[2pt]

CSFM{\scriptsize~\citep{gu2026cardiac}}
 & C
& 65.93 & 0.389 & 0.664
& 64.73 & 0.353 & 0.500
& 77.73 & 0.323 & 0.719 \\

SleepFM{\scriptsize~\citep{thapa2026sleepfm}}
 & Full
& 57.21 & 0.505 & 0.747
& 76.51 & 0.714 & 0.820
& 87.12 & 0.627 & 0.856 \\

SleepGPT{\scriptsize~\citep{huang2026sleepgpt}} & Full
& 72.74 & 0.635 & 0.804
& 78.34 & 0.776 & 0.870
& 90.73 & 0.832 & 0.935 \\

SleepMaMi{\scriptsize~\citep{park2026sleepmami}} & Full
& 83.07 & 0.732 & 0.831
& 80.95 & \textbf{0.798} & \textbf{0.882}
& 88.67 & \underline{0.841} & 0.937 \\

OSF{\scriptsize~\citep{shuai2026osf}} & Full
& 80.01 & 0.715 & 0.856
& 80.38 & 0.771 & 0.868
& 90.29 & 0.775 & 0.931 \\

\addlinespace[2pt]
\rowcolor{FTPurple}
SleepXA (Ours) & Full
& \textbf{84.25} & \textbf{0.772} & \textbf{0.884}
& \textbf{81.10} & 0.791 & \textbf{0.882}
& \underline{91.07} & 0.839 & 0.936 \\

\bottomrule[1.0pt]
\end{tabularx}

\par\vspace{3pt}
\begin{minipage}{\linewidth}
\scriptsize
C: ECG; Full: full experimental PSG configuration. AUP and AUR denote AUPRC
and AUROC, respectively. Respiratory-event detection is performed on 30-s epochs; an epoch is labeled
positive if at least 10~s of the epoch overlaps with an annotated apnea or
hypopnea event.
\end{minipage}
\vspace{-10pt} 
\endgroup
\end{table}

\paragraph{Sleep Spindle Detection.}
\label{absec:spindle_detection}
We evaluate single-channel spindle detection on MASS-SS2 using C3 EEG
and report FT results only,
following the
 SleepGPT evaluation protocol~\citep{huang2026sleepgpt}.
As shown in Table~\ref{tab:spindle_detection}, SleepXA achieves the highest
precision among the evaluated foundation-model methods for both expert
annotations and obtains F1 scores of 0.649 and 0.757 for E1 and E2,
respectively.
Because SleepGPT explicitly models time--frequency structure during pre-training,
it provides a particularly relevant comparison for spindle detection.
SleepXA improves F1 over SleepGPT by 3.9 and 1.6 percentage points for
E1 and E2, respectively. Compared with task-specific detectors, SleepXA surpasses SUMO~\citep{kaulen2022advanced} in F1 despite not being specifically designed for spindle detection, while remaining below RED-Time~\citep{tapia2020red} due to its comparatively higher precision but lower recall.

\begin{table}[t]
\centering
\caption{
\textbf{Sleep spindle detection on MASS-SS2 using C3 EEG.}
E1 and E2 denote the two expert annotations.
\textbf{Bold} and \underline{underlined} values indicate the
\textbf{best} and \underline{second-best} results across all methods
in each column, respectively, based on unrounded scores.
Higher is better for all metrics.
}
\label{tab:spindle_detection}
\begingroup
\footnotesize
\setlength{\tabcolsep}{3.0pt}
\renewcommand{\arraystretch}{0.92}

\begin{tabularx}{0.97\linewidth}{
@{}l*{6}{C}@{}
}
\toprule[1.0pt]

\multirow{2}{*}{\textbf{Model}}
& \multicolumn{3}{c}{\textbf{E1}}
& \multicolumn{3}{c}{\textbf{E2}} \\

\cmidrule(lr){2-4}
\cmidrule(lr){5-7}

& \textbf{Precision}
& \textbf{Recall}
& \textbf{F1}
& \textbf{Precision}
& \textbf{Recall}
& \textbf{F1} \\

\midrule
\rowcolor{SectionGray}
\multicolumn{7}{l}{
            \textit{Task-specific supervised models}
        } \\
        RED-Time{\scriptsize~\citep{tapia2020red}}

        & 0.583 & \textbf{0.866} & \textbf{0.681}
        & \underline{0.744} & \textbf{0.841} & \textbf{0.783} \\
        SUMO{\scriptsize~\citep{kaulen2022advanced}}
        & 0.527 & \underline{0.864} & 0.641
        & 0.694 & 0.807 & 0.726 \\

\midrule
\rowcolor{FTPurple}
\multicolumn{7}{@{}l}{
\textit{Foundation-model-based methods: full fine-tuning (FT)}
} \\
\addlinespace[0.5pt]
        SleepFM{\scriptsize~\citep{thapa2026sleepfm}}
        & 0.529 & 0.781 & 0.621
        & 0.702 & 0.801 & 0.741 \\
        SleepGPT{\scriptsize~\citep{huang2026sleepgpt}}
        & \underline{0.673} & 0.603 & 0.610
        & 0.702 & 0.801 & 0.741 \\
        
        SleepMaMi{\scriptsize~\citep{park2026sleepmami}} 
        & 0.541 & 0.729 & 0.551
        & 0.683 & 0.723 & 0.672 \\

        OSF{\scriptsize~\citep{shuai2026osf}}
        & 0.544 & 0.853 & 0.648
        & 0.710 & \underline{0.824} & 0.756 \\
        \rowcolor{FTPurple}
        SleepXA (ours)
        & \textbf{0.679} & 0.724 & \underline{0.649}
        & \textbf{0.782} & 0.742 & \underline{0.757} \\

\bottomrule[1.0pt]

\end{tabularx}
\begin{minipage}{\linewidth}
\scriptsize
Precision, recall, and F1 are averaged independently across folds.
\end{minipage}
\endgroup
\end{table}

\subsubsection{Detailed Clinical and Disease Evaluation}
\label{app:clinical_results}

This section provides additional task definitions and evaluation details for
the clinical benchmarks summarized in the main text.
The evaluations cover OSA severity assessment, heterogeneous sleep-disorder
classification, current disease diagnosis, and longitudinal clinical outcome
prediction.

\begin{table}[t]
\centering
\caption{
\textbf{Performance comparison on AHI estimation and OSA severity classification
across two representative cohorts.}
}
\label{tab:osa_severity_results}

\begingroup
\footnotesize
\setlength{\tabcolsep}{2.5pt}
\renewcommand{\arraystretch}{0.92}

\resizebox{0.97\linewidth}{!}{%
\begin{tabular}{@{}lc*{10}{c}@{}}
\toprule[1.0pt]

\multirow{2}{*}{\textbf{Model}}
& \multirow{2}{*}{\textbf{Channels}}
& \multicolumn{5}{c}{\textbf{Test SHHS1/2}}
& \multicolumn{5}{c}{\textbf{WSC}} \\

\cmidrule(lr){3-7}
\cmidrule(lr){8-12}

& &
\textbf{MAE} $\downarrow$
& \textbf{RMSE} $\downarrow$
& $\mathbf{R^2}$ $\uparrow$
& \textbf{Acc. (\%)} $\uparrow$
& \textbf{MF1 (\%)} $\uparrow$
& \textbf{MAE} $\downarrow$
& \textbf{RMSE} $\downarrow$
& $\mathbf{R^2}$ $\uparrow$
& \textbf{Acc. (\%)} $\uparrow$
& \textbf{MF1 (\%)} $\uparrow$ \\

\midrule
\rowcolor{SectionGray}
\multicolumn{12}{@{}l}{
\textit{General non-foundation models}
} \\
\addlinespace[0.5pt]

TimesNet{\scriptsize~\citep{wu2022timesnet}} & Full
& 5.319 & 7.424 & 0.751 & 66.50 & 64.36
& 3.276 & \textbf{5.136} & \textbf{0.911} & 78.31 & 75.47 \\

CNN-Transformer{\scriptsize~\citep{peh2022transformer}} & Full
& 10.473 & 15.436 & -0.076 & 39.50 & 14.16
& 10.317 & 17.915 & -0.088 & 27.71 & 10.85 \\

ResNet1D-50{\scriptsize~\citep{ribeiro2020automatic}} & Full
& 4.833 & 7.197 & 0.766 & 69.50 & 69.73
& 4.461 & 8.467 & 0.757 & 75.10 & 69.98 \\

ST-Transformer{\scriptsize~\citep{song2021transformer}} & Full
& 4.746 & 6.893 & 0.785 & 67.50 & 67.54
& 4.546 & 6.950 & 0.836 & 69.88 & 67.26 \\

ContraWR{\scriptsize~\citep{yang2023self}} & Full
& 4.829 & 7.253 & 0.763 & 66.00 & 64.31
& 3.359 & 6.542 & 0.855 & 77.91 & 73.14 \\

iTransformer{\scriptsize~\citep{liu2024itransformer}} & Full
& 6.192 & 8.552 & 0.670 & 59.00 & 57.24
& 4.499 & 8.536 & 0.753 & 74.70 & 70.01 \\

\midrule
\rowcolor{SectionGray}
\multicolumn{12}{@{}l}{
\textit{Task-specific supervised models}
} \\
\addlinespace[0.5pt]

TSDNet{\scriptsize~\citep{hu2025transparent}} & Full
& \underline{4.363} & \underline{6.171} & \underline{0.828} & 70.00 & 69.50
& \underline{3.234} & 6.116 & 0.873 & \underline{80.32} & 75.64 \\

\midrule
\rowcolor{LPBlue}
\multicolumn{12}{@{}l}{
\textit{Foundation models: linear probing (LP)}
} \\
\addlinespace[0.5pt]

CSFM{\scriptsize~\citep{gu2026cardiac}} & C
& 10.531 & 15.008 & -0.017 & 35.50 & 19.75
& 9.935 & 17.155 & 0.003 & 35.34 & 23.93 \\

SleepFM{\scriptsize~\citep{thapa2026sleepfm}} & Full
& 6.085 & 8.781 & 0.652 & 56.50 & 45.98
& 4.349 & 7.504 & 0.809 & 67.47 & 63.89 \\

SleepGPT{\scriptsize~\citep{huang2026sleepgpt}} & Full
& 5.692 & 8.088 & 0.705 & 59.00 & 53.37
& 4.260 & 7.082 & 0.830 & 71.89 & 66.72 \\

SleepMaMi{\scriptsize~\citep{park2026sleepmami}}  & Full
& 4.592 & 6.774 & 0.793 & 69.00 & 67.23
& 4.772 & 6.889 & 0.839 & 70.68 & 70.51 \\

OSF{\scriptsize~\citep{shuai2026osf}} & Full
& 5.239 & 7.235 & 0.764 & 66.50 & 65.34
& 4.571 & 7.354 & 0.817 & 69.88 & 67.29 \\

\addlinespace[2pt]
\rowcolor{LPBlue}
\textbf{SleepXA (Ours)} & Full
& 4.526 & 6.671 & 0.799 & 68.50 & 68.35
& 3.441 & 6.341 & 0.864 & 79.12 & 75.05 \\

\midrule
\rowcolor{FTPurple}
\multicolumn{12}{@{}l}{
\textit{Foundation models: full fine-tuning (FT)}
} \\
\addlinespace[0.5pt]

CSFM{\scriptsize~\citep{gu2026cardiac}} & C
& 10.407 & 15.312 & -0.059 & 36.00 & 20.23
& 9.641 & 16.885 & 0.034 & 38.55 & 25.84 \\

SleepFM{\scriptsize~\citep{thapa2026sleepfm}} & Full
& 4.846 & 6.984 & 0.780 & 59.50 & 59.21
& 4.178 & 7.325 & 0.818 & 74.71 & 67.35 \\

SleepGPT{\scriptsize~\citep{huang2026sleepgpt}}
 & Full
& 4.596 & 6.751 & 0.794 & 69.00 & 68.37
& 3.731 & \underline{5.464} & \underline{0.899} & 79.14 & \underline{76.28} \\

SleepMaMi{\scriptsize~\citep{park2026sleepmami}}  & Full
& 4.442 & 6.496 & 0.809 & \underline{70.50} & \underline{70.51}
& 3.363 & 6.248 & 0.868 & 78.31 & 74.95 \\

OSF{\scriptsize~\citep{shuai2026osf}} & Full
& 4.694 & 6.424 & 0.814 & 63.00 & 60.61
& 3.731 & 5.948 & 0.880 & 77.51 & 72.69 \\

\addlinespace[2pt]
\rowcolor{FTPurple}
\textbf{SleepXA (Ours)} & Full
& \textbf{4.200} & \textbf{5.920} & \textbf{0.842} & \textbf{73.50} & \textbf{73.32}
& \textbf{3.120} & 6.008 & 0.878 & \textbf{82.73} & \textbf{78.46} \\

\bottomrule[1.0pt]
\end{tabular}%
} %

\par\vspace{3pt}
\begin{minipage}{\linewidth}
\scriptsize
OSA severity is categorized into normal, mild, moderate, and severe groups
according to the reference AHI.
\end{minipage}

\endgroup
\end{table}

\paragraph{AHI Estimation and OSA Severity Classification.}
We evaluate continuous apnea--hypopnea index (AHI) estimation together with
four-level obstructive sleep apnea (OSA) severity classification on
Test SHHS1/2 and WSC.
OSA severity is defined using the conventional AHI thresholds:
normal ($<5$ events/h), mild ($5$--$<15$), moderate ($15$--$<30$),
and severe ($\geq30$ events/h)~\citep{berry2017aasm}.
Predicted severity is obtained by applying the same thresholds to the
estimated AHI.
This joint evaluation therefore assesses both continuous estimation of
respiratory burden and clinically defined severity stratification. The complete comparison with task-specific supervised models and foundation
models under LP and FT is reported in
Table~\ref{tab:osa_severity_results}. SleepGPT, SleepMaMi, and OSF also provide competitive
subject-level predictions, although SleepXA achieves lower AHI MAE
and higher severity Macro-F1 than these models on both cohorts
under both LP and FT.
The results complement the epoch-level respiratory-event evaluation by
examining whether the learned representations can support subject-level
estimation of overnight disease burden and clinically meaningful severity.

\begin{figure*}[t]
    \centering
    \includegraphics[
        width=\textwidth
    ]{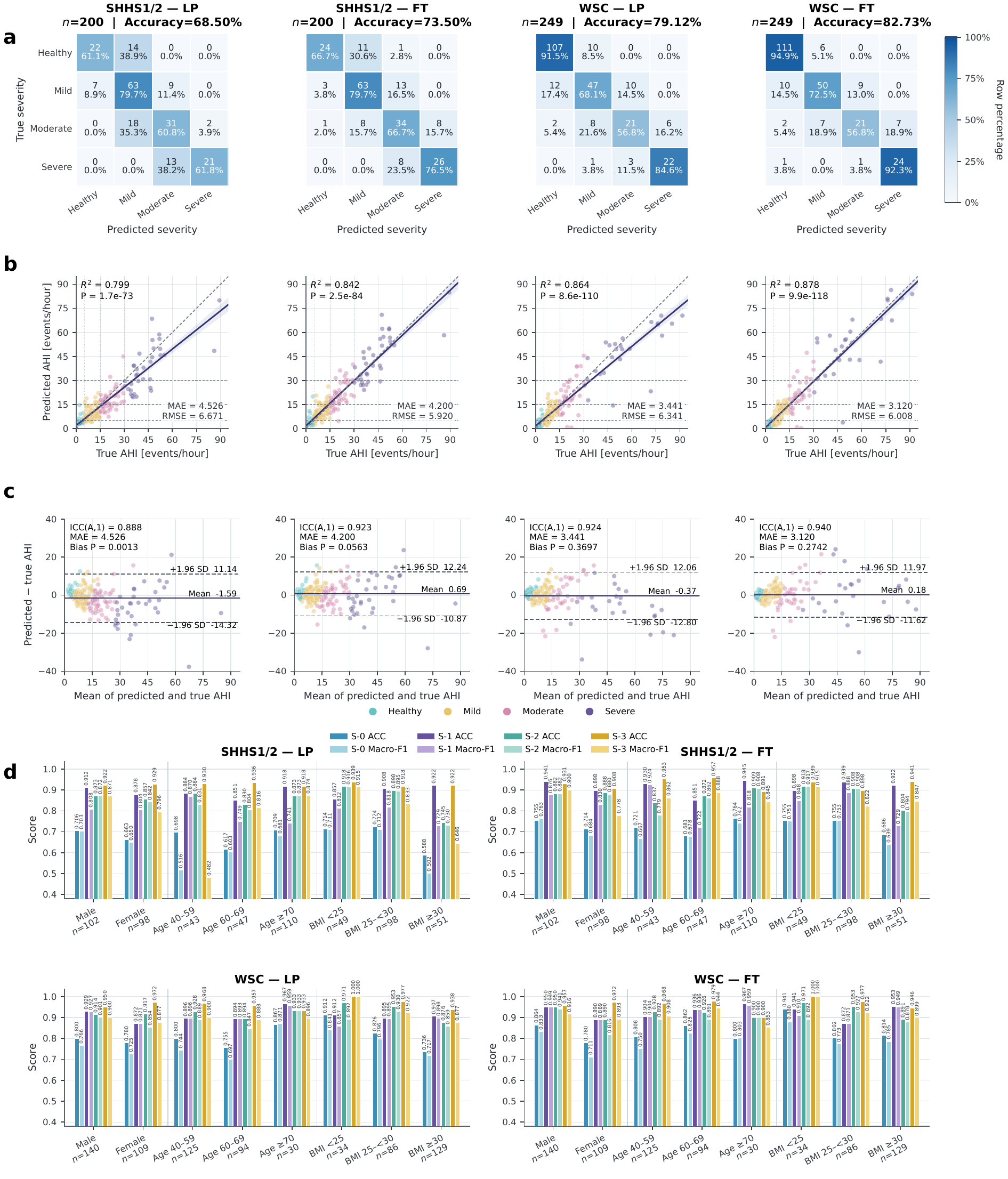}
    \vspace{-10pt}
    \caption{
    \textbf{Detailed analysis of AHI estimation and OSA severity prediction.}
    Results are shown for Test SHHS1/2 and WSC under linear probing (LP) and
    full fine-tuning (FT).
    (a) Four-level OSA severity confusion matrices for normal, mild, moderate,
    and severe OSA;
    (b) predicted versus reference AHI with regression statistics;
    (c) Bland--Altman agreement analysis between predicted and reference AHI;
    and
    d) subgroup performance across sex, age, and BMI strata, with S-0 to S-3 denoting normal, mild, moderate, and severe OSA, respectively. The subgroup analysis is included as a robustness diagnostic to examine
    whether predictive performance is concentrated in particular demographic
    groups.
    }
    \label{fig:ahi_severity_detailed}
    \vspace{-10pt}
\end{figure*}

\paragraph{Detailed OSA Prediction Analysis.}
Beyond aggregate metrics, we further analyze the prediction behavior of
SleepXA on Test SHHS1/2 and WSC under both LP and FT.
Figure~\ref{fig:ahi_severity_detailed} provides four complementary views.
First, the four-level confusion matrices characterize the pattern of errors
among normal, mild, moderate, and severe OSA.
Second, predicted-versus-reference AHI plots examine the correspondence
between estimated and observed respiratory burden across its continuous
range.
Third, Bland--Altman analyses provide an additional view of agreement and
systematic prediction bias. We additionally examine performance across sex, age, and BMI subgroups.
This analysis is particularly relevant because demographic information,
including age, is itself represented in the learned sleep representation,
as demonstrated by the demographic-prediction experiments.
The subgroup analysis therefore serves as a diagnostic robustness check for
whether OSA performance is concentrated in a narrow demographic stratum.
Across the considered subgroups, the model retains useful severity
classification performance under both LP and FT.
This subgroup analysis is intended to characterize how performance varies
across clinically relevant participant groups.

\vspace{-5pt}
\paragraph{Heterogeneous Sleep-Disorder Generalization.}
\label{app:cap_generalization}

We further evaluate whether the learned representations transfer beyond
sleep-disordered breathing to heterogeneous sleep pathologies using the
CAP Sleep Database~\citep{terzano2001atlas}.
The cohort includes multiple disorder categories, including nocturnal frontal
lobe epilepsy (NFLE), REM sleep behavior disorder (RBD), periodic leg
movements (PLM), insomnia, narcolepsy, and other sleep disorders.
We formulate a subject-level binary classification task that distinguishes
participants with sleep disorders from healthy controls and use subject-level
five-fold cross-validation because of the limited cohort size.

\begin{figure}[t]
    \centering
    \includegraphics[width=\linewidth]{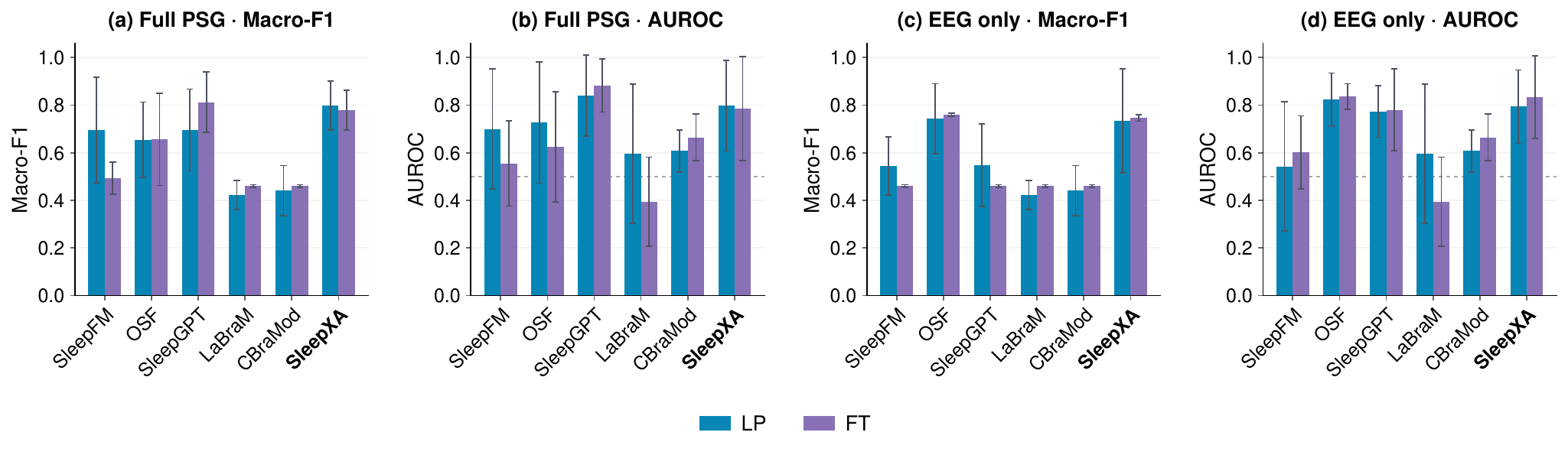}
    \vspace{-20pt}
    \caption{
    \textbf{Sleep-disorder classification on the CAP Sleep Database.}
    Subject-level binary classification of heterogeneous sleep disorders versus
    healthy controls under full PSG (a--b) and central EEG-only (c--d)
    configurations.
    Results are reported as mean and standard deviation across five
    subject-level folds.
    }
    \label{fig:cap_disorder_classification}
    \vspace{-8pt}
\end{figure}

We evaluate two sensing configurations: full multimodal PSG and a unified
central-EEG setting.
The full-PSG setting examines whether the multisystem representation transfers
to heterogeneous pathological conditions, whereas the EEG-only setting tests
whether disease-related information remains accessible under substantially
restricted sensing.
As shown in Fig.~\ref{fig:cap_disorder_classification}, SleepXA maintains
effective transfer across both settings.
With full PSG, the frozen representation reaches 
0.798 Macro-F1, indicating that disorder-related information is already
accessible without adapting the pretrained backbone.
The central-EEG setting provides a complementary test of transfer when only a
restricted subset of the original PSG signals is available.

\subsubsection{Current and Longitudinal Clinical Prediction}
\label{app:clinical_prediction}

\paragraph{Task definitions.}
We consider 12 subject-level clinical prediction tasks spanning current disease
status and longitudinal outcomes.
Current disease diagnosis includes six conditions across three clinical
domains: respiratory disorders (COPD and emphysema), mental-health conditions
(clinical depression and depressive affect), and cardiovascular conditions
(heart failure and stroke)~\citep{quan1997sleep,hou2026omni}.
Whole-night recordings from Test SHHS1 and Test SHHS2 are used for these
evaluations, with all recordings from the same participant assigned to the
same fold.

Longitudinal sleep prediction uses Test SHHS1 recordings to predict AHI
worsening and incident hypersomnia observed at the Test SHHS2 follow-up visit. We further use Test SHHS2 as the index examination to predict four five-year cardiovascular outcomes: a composite CVD endpoint and three specific outcomes, namely heart failure, stroke, and cardiovascular death. Participants without sufficient follow-up for the corresponding prediction horizon are excluded. The composite CVD endpoint is defined as the occurrence of angina, heart failure, myocardial infarction, stroke, coronary revascularization, or cardiovascular death within five years after SHHS2.

 \vspace{-8pt}
\paragraph{Evaluation protocol.}
Given the limited number of positive participants and substantial class
imbalance in several endpoints, we evaluate these tasks using frozen
pretrained representations.
Epoch-level representations are aggregated over each overnight recording using their mean and standard deviation, followed by feature standardization and an $\ell_2$-regularized logistic-regression probe. Within each cross-validation fold, model hyperparameters are selected using validation data drawn only from the training partition, while the held-out fold remains untouched for final evaluation. We use participant-level five-fold cross-validation, with paired SHHS1 and SHHS2 visits from the same participant retained in the same fold.

\vspace{-8pt}
\paragraph{Current disease diagnosis.}
\label{Current disease diagnosis.}
Table~\ref{tab:current_disease_results} reports frozen-representation
performance for the six current disease diagnoses.
SleepXA achieves the highest mean AUROC among the evaluated foundation models
on five of the six tasks, spanning respiratory, mental-health, and
cardiovascular conditions.
The only exception is depressive affect, for which CBraMod achieves the
highest mean AUROC.

\begin{table*}[t]
    \centering
    \caption{
    \textbf{Linear-probing performance for current disease diagnosis.}
 }
    \label{tab:current_disease_results}

    \begingroup
    \footnotesize
    \setlength{\tabcolsep}{3.2pt}
    \renewcommand{\arraystretch}{1.08}

    \resizebox{0.96\textwidth}{!}{%
    \begin{tabular}{lcccccc}
        \toprule
        \multirow{2}{*}{\textbf{Method}}
        & \multicolumn{2}{c}{\textbf{Respiratory}}
        & \multicolumn{2}{c}{\textbf{Mental Health}}
        & \multicolumn{2}{c}{\textbf{Cardiovascular}} \\
        \cmidrule(lr){2-3}
        \cmidrule(lr){4-5}
        \cmidrule(lr){6-7}

        & COPD
        & Emphysema
        & Clin. Depression
        & Depressive Affect
        & Heart Failure
        & Stroke \\
        \midrule
        
        LaBraM{\scriptsize~\citep{jiang2024large}}
        & 0.469$\pm$0.061
        & 0.553$\pm$0.025
        & 0.487$\pm$0.034
        & 0.561$\pm$0.077
        & 0.599$\pm$0.051
        & 0.573$\pm$0.051 \\

        CBraMod{\scriptsize~\citep{wang2025cbramod}}
        & 0.502$\pm$0.120
        & 0.548$\pm$0.054
        & 0.502$\pm$0.037
        & \textbf{0.629$\pm$0.064}
        & 0.618$\pm$0.048
        & 0.582$\pm$0.043 \\

        SleepFM{\scriptsize~\citep{thapa2026sleepfm}}
        & 0.553$\pm$0.087
        & 0.596$\pm$0.097
        & 0.452$\pm$0.021
        & 0.544$\pm$0.064
        & 0.704$\pm$0.073
        & 0.555$\pm$0.028 \\

        SleepGPT{\scriptsize~\citep{huang2026sleepgpt}}
        & 0.549$\pm$0.052
        & 0.566$\pm$0.076
        & 0.520$\pm$0.044
        & 0.580$\pm$0.073
        & 0.643$\pm$0.050
        & 0.524$\pm$0.051 \\

        SleepMaMi{\scriptsize~\citep{park2026sleepmami}}
        & 0.627$\pm$0.058
        & 0.575$\pm$0.050
        & 0.542$\pm$0.044
        & 0.552$\pm$0.089
        & 0.702$\pm$0.055
        & 0.480$\pm$0.032 \\

       OSF{\scriptsize~\citep{shuai2026osf}}
        & 0.642$\pm$0.055
        & 0.636$\pm$0.044
        & 0.535$\pm$0.028
        & 0.593$\pm$0.068
        & 0.679$\pm$0.039
        & 0.545$\pm$0.026 \\

        \rowcolor{LPBlue}
        SleepXA (ours)
        & \textbf{0.682$\pm$0.046}
        & \textbf{0.678$\pm$0.046}
        & \textbf{0.547$\pm$0.031}
        & 0.591$\pm$0.069
        & \textbf{0.709$\pm$0.057}
        & \textbf{0.604$\pm$0.026} \\

        \bottomrule
    \end{tabular}%
    }

    \endgroup
\end{table*}

\vspace{-10pt}

\paragraph{Longitudinal outcome prediction.}
Table~\ref{tab:longitudinal_disease_results} reports results for two
cross-visit sleep outcomes and four five-year cardiovascular outcomes.
SleepXA achieves the highest mean AUROC among the evaluated foundation models
on all six longitudinal tasks.
These results complement the current-disease evaluation by testing whether
information retained in the frozen representation is also relevant to
subsequent changes in sleep-related phenotypes and longer-term cardiovascular
outcomes.

\begin{table*}[t]
    \centering
    \caption{
    \textbf{Longitudinal sleep-related progression and five-year cardiovascular outcome prediction.}
    Longitudinal sleep outcomes use SHHS1 representations to predict changes observed at
    SHHS2; the cardiovascular outcomes use SHHS2 as the index examination for
    five-year prediction.}
    \label{tab:longitudinal_disease_results}

    \begingroup
    \footnotesize
    \setlength{\tabcolsep}{2.8pt}
    \renewcommand{\arraystretch}{1.08}

    \resizebox{0.95\textwidth}{!}{%
    \begin{tabular}{lcccccc}
        \toprule
        \multirow{2}{*}{Method}
        & \multicolumn{2}{c}{Longitudinal Sleep Outcomes}
        & \multicolumn{4}{c}{5-Year Cardiovascular Outcomes} \\
        \cmidrule(lr){2-3}
        \cmidrule(lr){4-7}

        & AHI Worsening
        & Incident Hypersomnia
        & Composite CVD outcome
        & Heart Failure
        & Stroke
        & CVD Death \\
        \midrule

        LaBraM{\scriptsize~\citep{jiang2024large}}
        & 0.513$\pm$0.042
        & 0.493$\pm$0.107
        & 0.603$\pm$0.017
        & 0.648$\pm$0.048
        & 0.555$\pm$0.050
        & 0.675$\pm$0.051 \\

        CBraMod{\scriptsize~\citep{wang2025cbramod}}
        & 0.549$\pm$0.024
        & 0.556$\pm$0.070
        & 0.612$\pm$0.051
        & 0.665$\pm$0.053
        & 0.608$\pm$0.043
        & 0.669$\pm$0.076 \\

        SleepFM{\scriptsize~\citep{thapa2026sleepfm}}
        & 0.594$\pm$0.026
        & 0.559$\pm$0.060
        & 0.634$\pm$0.044
        & 0.695$\pm$0.046
        & 0.540$\pm$0.062
        & 0.695$\pm$0.032 \\

        SleepGPT{\scriptsize~\citep{huang2026sleepgpt}}
        & 0.567$\pm$0.036
        & 0.585$\pm$0.058
        & 0.605$\pm$0.025
        & 0.671$\pm$0.038
        & 0.516$\pm$0.074
        & 0.691$\pm$0.075 \\

        SleepMaMi{\scriptsize~\citep{park2026sleepmami}}
        & 0.595$\pm$0.026
        & 0.558$\pm$0.058
        & 0.640$\pm$0.021
        & 0.672$\pm$0.052
        & 0.507$\pm$0.038
        & 0.740$\pm$0.067 \\

        OSF{\scriptsize~\citep{shuai2026osf}}
        & 0.553$\pm$0.041
        & 0.552$\pm$0.047
        & 0.641$\pm$0.034
        & 0.692$\pm$0.040
        & 0.495$\pm$0.039
        & 0.740$\pm$0.038 \\
        
        \rowcolor{LPBlue}
        SleepXA (ours)
        & \textbf{0.614$\pm$0.055}
        & \textbf{0.595$\pm$0.110}
        & \textbf{0.643$\pm$0.014}
        & \textbf{0.706$\pm$0.032}
        & \textbf{0.613$\pm$0.048}
        & \textbf{0.757$\pm$0.068} \\

        \bottomrule
    \end{tabular}%
    }

    \endgroup
\end{table*}

\subsubsection{Discussion of Daytime Physiological Transfer}
\label{app:daytime_transfer_discussion}

The daytime transfer results show that representations learned from nocturnal
PSG can remain useful beyond sleep-specific tasks.
Sleep foundation models exhibit meaningful transfer to awake-state EEG tasks,
including MDD~\citep{mumtaz2017electroencephalogram},
Alzheimer's disease~\citep{miltiadous2023dataset},
mental-arithmetic recognition~\citep{zyma2019electroencephalograms},
motor imagery~\citep{tangermann2012review}, and vigilance estimation
~\citep{zheng2017multimodal}.
This suggests that \textbf{sleep pre-training captures neural structure that remains
relevant across behavioral states rather than being restricted to sleep-stage
information}.

SleepXA further shows strong transfer to tasks that differ more substantially
from nocturnal sleep, including ECG-based atrial-fibrillation recognition and
surface-EMG gesture classification.
The SEED-VIG evaluation provides a complementary test of multimodal transfer:
using EEG+EOG improves vigilance estimation over EEG alone, indicating that
cross-modal complementarity remains useful in an awake-state task.
These results are consistent with the compositional design of SleepXA, which
inherits modality-specific knowledge from pretrained physiological encoders
and organizes it through sleep-guided common-specific alignment.
The larger differences observed under LP further suggest that this advantage
is already present in the pretrained representation, while FT can partially
reduce differences between pre-training strategies.
Overall, the results support nocturnal PSG as a useful context for \textbf{learning
reusable multisystem physiological representations beyond sleep}.

\vspace{-8pt}
\subsubsection{Effect of Physiological Foundation-Model Initialization}
\label{app:fm_initialization}

\begin{figure*}[!ht]
    \centering
    \includegraphics[
        width=\textwidth,
        trim={0cm 0cm 0cm 0cm},
        clip
    ]{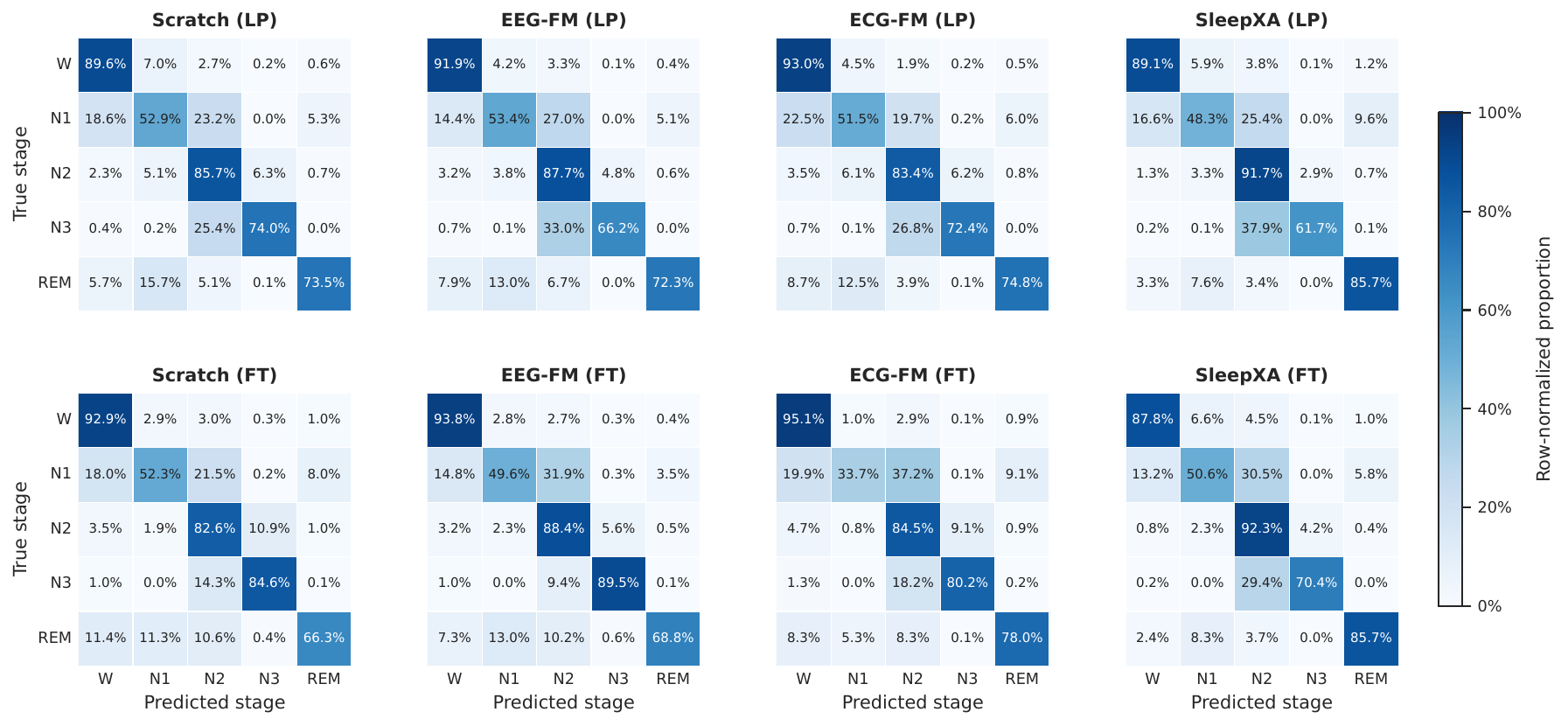}
    \caption{
    \textbf{Class-wise analysis of foundation-model initialization on sleep staging.}
        Row-normalized confusion matrices are shown for four initialization
    settings under linear probing (LP, top row) and full fine-tuning
    (FT, bottom row). Rows denote ground-truth stages and columns denote predicted stages. Each cell reports the percentage within the corresponding ground-truth class.
    }
    \label{fig:initialization_confusion_matrices}
    \vspace{-6pt}
\end{figure*}

We first examine whether the performance of SleepXA benefits from physiological
knowledge inherited from pretrained EEG and ECG foundation models.
All variants use the same SleepXA architecture, multimodal inputs, alignment
objectives, pre-training data, downstream splits, and optimization settings,
differing only in EEG and ECG encoder initialization.
\textit{Scratch} randomly initializes both encoders while keeping the remaining architecture and training configuration unchanged;
\textit{EEG-FM} and \textit{ECG-FM} initialize only the corresponding encoder
from LaBraM and CSFM, respectively; and
\textit{EEG+ECG-FM} initializes both encoders from their pretrained
checkpoints.
This controlled comparison isolates the contribution of inherited
physiological knowledge.
\vspace{-5pt}
\paragraph{Sleep staging.}
Table~\ref{tab:initialization_sleep_staging} reports the complete WSC
sleep staging results.
Under LP, joint EEG+ECG foundation-model initialization improves accuracy from
81.40\% with Scratch to 85.16\%, MF1 from 73.00\% to 76.09\%, and
Cohen's $\kappa$ from 0.717 to 0.765.
The same trend remains after FT, where the jointly initialized model reaches
85.95\% accuracy, 77.61\% MF1, and $\kappa=0.780$.
The class-wise confusion matrices in
Fig.~\ref{fig:initialization_confusion_matrices} further show improvements
across multiple sleep stages rather than only the dominant N2 class.
Notably, ECG initialization improves REM recognition, consistent with the
distinct autonomic and cardiac dynamics observed during REM sleep~\citep{berry2017aasm}.

\begin{table*}[t]
\centering
\caption{
\textbf{Controlled initialization study on WSC sleep staging.}
All variants share the same architecture and training configuration,
differing only in the initialization of the EEG and ECG encoders.
All metrics except Cohen's $\kappa$ are reported in percentages.
\textbf{Bold} and \underline{underlined} values denote the
\textbf{best} and \underline{second-best} results within LP or FT,
respectively.
}
\label{tab:initialization_sleep_staging}

\vspace{1mm}

\begingroup
\footnotesize
\setlength{\tabcolsep}{2.8pt}
\renewcommand{\arraystretch}{0.92}

\begin{tabularx}{0.97\linewidth}{
@{}lc*{8}{C}@{}
}
\toprule[1.0pt]

\textbf{Protocol}
& \textbf{Initialization}
& \textbf{Acc.}
& \textbf{MF1}
& $\boldsymbol{\kappa}$
& {W F1}
& {N1 F1}
& {N2 F1}
& {N3 F1}
& \mbox{REM F1} \\

\midrule

\multirow{4}{*}{\textbf{LP}}
& Scratch
& 81.40
& 73.00
& 0.717
& 85.61
& 46.07
& 88.34
& \underline{63.33}
& 81.63 \\

& EEG-FM
& \underline{82.36}
& \underline{73.79}
& \underline{0.727}
& \underline{85.89}
& \textbf{50.75}
& \underline{88.58}
& 62.64
& 81.11 \\

& ECG-FM
& 80.77
& 72.49
& 0.710
& 84.38
& 46.05
& 87.44
& 62.44
& \underline{82.16} \\

\addlinespace[2pt]
\rowcolor{LPBlue}
& \textbf{SleepXA}
& \textbf{85.16}
& \textbf{76.09}
& \textbf{0.765}
& \textbf{87.27}
& \underline{49.32}
& \textbf{91.04}
& \textbf{65.28}
& \textbf{87.52} \\

\midrule

\multirow{4}{*}{\textbf{FT}}
& Scratch
& 79.93
& 71.84
& 0.698
& 84.28
& \textbf{54.76}
& 86.54
& 58.46
& 75.17 \\

& EEG-FM
& \underline{83.67}
& \underline{76.04}
& \underline{0.748}
& \underline{86.91}
& 51.27
& \underline{89.52}
& \textbf{73.20}
& 79.30 \\

& ECG-FM
& 81.19
& 71.83
& 0.712
& 84.62
& 44.82
& 86.68
& 60.23
& \underline{82.82} \\

\addlinespace[2pt]
\rowcolor{FTPurple}
& \textbf{SleepXA}
& \textbf{85.95}
& \textbf{77.61}
& \textbf{0.780}
& \textbf{88.73}
& \underline{51.81}
& \textbf{91.23}
& \underline{67.02}
& \textbf{89.25} \\

\bottomrule[1.0pt]

\end{tabularx}

\endgroup
\end{table*}

\begin{table}[t]
\centering
\caption{
\textbf{Controlled initialization study on the CinC2017 AF task.}
Results are reported as mean $\pm$ standard deviation across test folds to assess
the stability of different initialization strategies.
\textbf{Bold} and \underline{underlined} values denote the
\textbf{best} and \underline{second-best} results within LP or FT,
respectively.
}
\label{tab:initialization_cinc2017}

\begingroup
\footnotesize
\setlength{\tabcolsep}{4.5pt}
\renewcommand{\arraystretch}{0.92}

\resizebox{0.65\linewidth}{!}{%
\begin{tabular}{@{}llccc@{}}
\toprule[1.0pt]

\textbf{Protocol}
& \textbf{Initialization}
& \textbf{Ch-F1 $\uparrow$}
& \textbf{AF-F1 $\uparrow$}
& \textbf{AF-AUROC $\uparrow$} \\

\midrule

\multirow{4}{*}{\textbf{LP}}
& Scratch
& $0.346 \pm 0.005$
& $0.181 \pm 0.035$
& $0.603 \pm 0.033$ \\

& EEG-FM
& $0.349 \pm 0.011$
& $0.129 \pm 0.034$
& $0.602 \pm 0.039$ \\

& ECG-FM
& $\underline{0.569 \pm 0.027}$
& $\underline{0.398 \pm 0.030}$
& $\underline{0.808 \pm 0.033}$ \\

\rowcolor{LPBlue}
& \textbf{SleepXA}
& $\mathbf{0.732 \pm 0.004}$
& $\mathbf{0.743 \pm 0.019}$
& $\mathbf{0.912 \pm 0.027}$ \\

\midrule

\multirow{4}{*}{\textbf{FT}}
& Scratch
& $0.446 \pm 0.020$
& $0.396 \pm 0.026$
& $0.846 \pm 0.015$ \\

& EEG-FM
& $0.475 \pm 0.040$
& $0.415 \pm 0.050$
& $0.856 \pm 0.022$ \\

& ECG-FM
& $\underline{0.812 \pm 0.010}$
& $\underline{0.808 \pm 0.014}$
& $\mathbf{0.984 \pm 0.002}$ \\

\rowcolor{FTPurple}
& \textbf{SleepXA}
& $\mathbf{0.834 \pm 0.013}$
& $\mathbf{0.823 \pm 0.022}$
& $\underline{0.984 \pm 0.003}$ \\

\bottomrule[1.0pt]

\end{tabular}%
}

\endgroup
\end{table}

\paragraph{Cross-context ECG transfer.}
We further evaluate the same initialization variants on CinC2017
atrial-fibrillation recognition, transferring representations learned from
nocturnal multimodal PSG to an independent single-lead ECG task.
As shown in Table~\ref{tab:initialization_cinc2017}, ECG-FM initialization
already improves AF-F1 from 0.181 with Scratch to 0.398 under LP.
Joint EEG+ECG initialization further increases AF-F1 to 0.743 and
Ch-F1 to 0.732.
After FT, the differences become smaller, although the jointly initialized
model retains the highest Ch-F1 and AF-F1.

The pronounced LP improvement is particularly informative because the
pretrained representation remains frozen during downstream adaptation.
These results indicate that PSG-based composition preserves the
modality-specific cardiac knowledge inherited from the ECG foundation model,
while integrating it with complementary neural representations.

\subsubsection{Robustness to Physiological Foundation-Model Backbones}
\label{app:backbone_robustness}

Building on the controlled initialization study above, we examine whether the
benefit of pretrained physiological initialization generalizes across different
constituent foundation models.
We consider LaBraM~\citep{jiang2024large} and CBraMod~\citep{wang2025cbramod} as EEG backbones and CSFM~\citep{gu2026cardiac} and ECGFounder~\citep{li2025electrocardiogram} as ECG backbones,
resulting in four EEG--ECG backbone combinations.
For each pair, pretrained initialization is compared with an
architecture-matched Scratch counterpart under the same pre-training and
evaluation protocol.
All experiments use WSC sleep staging under LP with the same fixed random seed.

As shown in Table~\ref{tab:backbone_robustness_full}, pretrained initialization
improves over Scratch for all four backbone combinations.
The gains range from 1.98 to 6.12 percentage points in accuracy and from
3.09 to 11.24 points in MF1.
The default LaBraM+CSFM configuration reaches 85.16\% accuracy,
76.09\% MF1, and $\kappa=0.765$, while the alternative backbone
combinations also benefit consistently from pretrained initialization. The class-wise results show that these gains are not confined to the dominant
N2 stage.
For example, with CBraMod+ECGFounder, N1 F1 increases from 28.02\% to
47.11\% and REM F1 from 60.90\% to 85.35\%, compared with a smaller change
in N2 F1 from 86.01\% to 88.51\%.
Pronounced improvements in REM recognition are also observed across the other
backbone combinations.

These consistent gains across heterogeneous EEG and ECG backbone pairs
show that SleepXA is not tied to the default LaBraM+CSFM configuration,
although the absolute performance still depends on the quality and
compatibility of the constituent pretrained representations.

\begin{table*}[t]
\centering
\caption{
\textbf{Robustness to different EEG and ECG foundation-model backbones on WSC
sleep staging.}
$\Delta$ denotes the improvement of pretrained initialization over Scratch; bold indicates the best result within each setting.
}
\label{tab:backbone_robustness_full}

\begingroup
\scriptsize
\setlength{\tabcolsep}{2.5pt}
\renewcommand{\arraystretch}{1.08}

\resizebox{0.9\textwidth}{!}{%
\begin{tabular}{@{}lll*{10}{c}@{}}
\toprule
\multirow{2}{*}{\textbf{EEG}}
& \multirow{2}{*}{\textbf{ECG}}
& \multirow{2}{*}{\textbf{Initialization}}
& \multicolumn{3}{c}{\textbf{Overall performance}}
& \multicolumn{5}{c}{\textbf{Class-wise F1}}
& \multicolumn{2}{c}{\textbf{Improvement}} \\
\cmidrule(lr){4-6}
\cmidrule(lr){7-11}
\cmidrule(lr){12-13}
& &
& Acc. & MF1 & $\kappa$
& W & N1 & N2 & N3 & REM
& \textbf{$\Delta$ Acc.} & \textbf{$\Delta$ MF1} \\
\midrule

\multirow{2}{*}{LaBraM}
& \multirow{2}{*}{CSFM}
& Scratch
& \textbf{81.40}
& \textbf{73.00}
& \textbf{0.717}
& \textbf{85.61}
& \textbf{46.07}
& 88.34
& \textbf{63.33}
& \textbf{81.63}
& -- & -- \\

& & Pretrained
& \textbf{85.16}
& \textbf{76.09}
& \textbf{0.765}
& \textbf{87.27}
& \textbf{49.32}
& \textbf{91.04}
& \textbf{65.28}
& \textbf{87.52}
& +3.76 & +3.09 \\

\midrule

\multirow{2}{*}{LaBraM}
& \multirow{2}{*}{ECGFounder}
& Scratch
& 80.38
& 68.01
& 0.692
& 84.37
& 32.37
& \textbf{88.48}
& 61.59
& 73.27
& -- & -- \\

& & Pretrained
& 82.36
& 74.28
& 0.733
& 86.39
& 47.38
& 88.51
& 64.17
& 84.93
& +1.98 & +6.27 \\

\midrule

\multirow{2}{*}{CBraMod}
& \multirow{2}{*}{ECGFounder}
& Scratch
& 76.30
& 63.04
& 0.627
& 80.94
& 28.02
& 86.01
& 59.35
& 60.90
& -- & -- \\

& & Pretrained
& 82.42
& 74.28
& 0.735
& 86.63
& 47.11
& 88.51
& 63.79
& 85.35
& +6.12 & +11.24 \\

\midrule

\multirow{2}{*}{CBraMod}
& \multirow{2}{*}{CSFM}
& Scratch
& 76.51
& 62.30
& 0.625
& 81.21
& 27.60
& 86.29
& 56.37
& 60.02
& -- & -- \\

& & Pretrained
& 80.91
& 73.21
& 0.714
& 86.20
& 47.53
& 87.58
& 62.25
& 82.48
& +4.40 & +10.91 \\

\bottomrule
\end{tabular}%
}

\par\vspace{3pt}
\begin{minipage}{0.99\textwidth}
\scriptsize
\textit{Note.}
Scratch and Pretrained use identical backbone architectures and differ only in
whether the EEG and ECG encoders are initialized from their corresponding
foundation-model checkpoints.
LaBraM+CSFM is the default SleepXA backbone combination.
All downstream results are obtained under LP.
\end{minipage}
\vspace{-10pt}
\endgroup
\end{table*}

\subsubsection{Detailed Comparison of Multimodal Integration Strategies}
\label{app:integration_comparison}

We provide the complete LP and FT results for the multimodal integration
comparison on WSC sleep staging and CinC2017 atrial-fibrillation recognition.
The WSC experiments evaluate robustness across different PSG sensor
configurations, while CinC2017 examines whether the learned representations
transfer beyond the sleep domain to an independent ECG recognition task.

We compare SleepXA with representative alternatives spanning direct feature
fusion, attention-based interaction, contrastive alignment, and common-specific
representation learning.
Specifically, we consider concatenation followed by an MLP fusion module,
cross-attention fusion~\citep{tsai2019multimodal}, CLIP-style contrastive alignment~\citep{radford2021learning}, MindMix-style
alignment~\citep{liu2026mindmix}, and DecAlign-style alignment~\citep{qian2026decalign}.
All strategies produce a fixed-dimensional downstream representation and are
evaluated using the same prediction head and downstream protocol.
The physiological encoders, pre-training data, downstream splits, and
optimization settings are otherwise kept consistent.

\vspace{-5pt}
\paragraph{Sleep staging under different sensor configurations.}
We first evaluate the integration strategies on WSC sleep staging under three
input configurations.
\textit{Full} uses all seven PSG modalities considered in this study.
\textit{Core} uses EEG, EOG, and EMG; and
\textit{Wearable} is restricted to EEG only.
This comparison tests whether an integration strategy learned from multimodal
PSG remains effective as the available physiological inputs are progressively
restricted.

As shown in Table~\ref{tab:integration_wsc}, SleepXA achieves the strongest
overall performance across the three sensor configurations under both LP and
FT.
The advantage is particularly consistent under LP, where the pretrained
representation remains frozen.
For Full PSG, SleepXA reaches an MF1 of 76.09\%, compared with 75.47\%
for the strongest alternative.
The corresponding MF1 values are 75.08\% under the Core configuration
and 69.60\% under EEG-only input.
The advantage therefore persists as the available sensing configuration is
reduced, suggesting that the learned representation does not depend on access
to the complete PSG montage. After FT, SleepXA also achieves the highest MF1 across all
three configurations, although the gaps to several alternative integration
strategies become smaller.
This pattern indicates that the choice of multimodal alignment has its clearest
effect on the quality of the pretrained frozen representation, while downstream
end-to-end optimization can partially compensate for differences introduced
during multimodal pre-training.

\begin{table*}[t]
\centering
\caption{
\textbf{Comparison of integration strategies across input configurations on WSC sleep staging.}
\textbf{Bold} and \underline{underlined} values denote the
\textbf{best} and \underline{second-best} results within each configuration
and adaptation setting.
}

\label{tab:integration_wsc}

\vspace{1mm}

\begingroup
\footnotesize
\setlength{\tabcolsep}{2.5pt}
\renewcommand{\arraystretch}{0.92}

\begin{tabularx}{0.97\linewidth}{
@{}ll*{6}{C}@{}
}
\toprule[1.0pt]

\multirow{2}{*}{\textbf{Method}}
& \multirow{2}{*}{\textbf{Config.}}
& \multicolumn{3}{c}{\textbf{LP}}
& \multicolumn{3}{c}{\textbf{FT}} \\

\cmidrule(lr){3-5}
\cmidrule(lr){6-8}

&
& Acc.
& MF1
& $\boldsymbol{\kappa}$
& Acc.
& MF1
& $\boldsymbol{\kappa}$ \\

\midrule

\textbf{SleepXA (Ours)}
& Full
& \textbf{85.16} & \textbf{76.09} & \textbf{0.765}
& \textbf{85.95} & \textbf{77.61} & \textbf{0.780} \\

MindMix-style Alignment
& Full
& 83.43 & 74.87 & 0.744
& 83.86 & 76.18 & 0.754 \\

Cross-Attention Fusion
& Full
& 83.74 & 75.18 & 0.748
& 84.57 & \underline{76.63} & 0.763 \\

Concatenation + MLP Fusion
& Full
& 83.75 & 74.46 & 0.747
& \underline{85.16} & 76.57 & \underline{0.769} \\

DecAlign-style Alignment
& Full
& 83.56 & \underline{75.47} & 0.746
& 84.60 & 76.36 & 0.765 \\

CLIP-style Contrastive Alignment
& Full
& \underline{84.24} & 75.26 & \underline{0.756}
& 83.26 & 75.84 & 0.754 \\

\midrule

\textbf{SleepXA (Ours)}
& Core
& \textbf{84.46} & \textbf{75.08} & \textbf{0.755}
& \textbf{85.50} & \textbf{77.80} & \textbf{0.778} \\

MindMix-style Alignment
& Core
& 82.68 & 73.79 & \underline{0.740}
& 83.11 & 76.38 & 0.747 \\

Cross-Attention Fusion
& Core
& 83.09 & \underline{74.44} & 0.738
& 83.87 & 75.11 & 0.748 \\

Concatenation + MLP Fusion
& Core
& \underline{83.11} & 73.46 & 0.737
& \underline{84.71} & \underline{77.20} & \underline{0.767} \\

DecAlign-style Alignment
& Core
& 82.82 & 73.45 & 0.731
& 83.67 & 75.87 & 0.759 \\

CLIP-style Contrastive Alignment
& Core
& 82.89 & 73.87 & 0.732
& 82.99 & 74.80 & 0.741 \\

\midrule

\textbf{SleepXA (Ours)}
& Wearable
& \textbf{81.02} & \textbf{69.60} & \textbf{0.701}
& \textbf{83.41} & \textbf{75.07} & \textbf{0.747} \\

MindMix-style Alignment
& Wearable
& 78.18 & 66.85 & 0.663
& 82.00 & 72.71 & 0.721 \\

Cross-Attention Fusion
& Wearable
& 78.16 & 67.62 & 0.662
& \underline{82.37} & \underline{73.15} & \underline{0.729} \\

Concatenation + MLP Fusion
& Wearable
& 77.79 & 66.62 & 0.658
& 80.24 & 71.00 & 0.717 \\

DecAlign-style Alignment
& Wearable
& 79.98 & 69.03 & 0.694
& 81.18 & 72.15 & 0.724 \\

CLIP-style Contrastive Alignment
& Wearable
& \underline{80.19} & \underline{69.37} & \underline{0.697}
& 80.42 & 71.31 &0.717 \\
\bottomrule[1.0pt]
\end{tabularx}

\endgroup
\end{table*}

\vspace{-5pt}
\paragraph{Cross-context ECG transfer.}
We next evaluate the different multimodal fusion and alignment strategies on
the CinC2017 AF task (Table~\ref{tab:integration_cinc}). Unlike WSC sleep staging, this setting transfers
representations learned from nocturnal multimodal PSG to an independent
single-lead ECG task outside the sleep context, providing a more stringent
test of whether different integration strategies preserve transferable
modality-specific physiological information.

Under LP, SleepXA achieves the highest Ch-F1 (73.18\%) and AF-F1
(74.30\%), exceeding the strongest alternative by 12.38 and 12.92 percentage
points, respectively.
Cross-Attention obtains a higher AF-AUROC (93.92\% versus 91.15\% for
SleepXA), showing that the advantage of SleepXA is more pronounced for
F1-based rhythm discrimination than for AUROC. After FT, the differences among the integration strategies become
substantially smaller.
Cross-Attention achieves the highest Challenge-F1 and AF-F1, while SleepXA
remains close to the best result across all three metrics.
The contrast between LP and FT is informative: differences in multimodal
pre-training are most apparent when the learned representation is frozen,
whereas sufficiently strong task-specific supervision can substantially reduce
these differences during end-to-end adaptation. Taken together with the WSC results, these experiments suggest that the
conditional common-specific alignment used by SleepXA primarily improves the
quality and transferability of the pretrained representation.
Its benefits persist under reduced sensor availability within the sleep domain
and are especially apparent when transferring to a physiologically distinct
single-modality task.

\begin{table}[t]
\centering
\caption{
\textbf{Cross-context transfer on CinC2017 AF task.}
Results are reported as percentages.
Best and second-best results are shown in bold and underlined, respectively.
}
\label{tab:integration_cinc}
\begingroup
\small
\setlength{\tabcolsep}{3.0pt}
\renewcommand{\arraystretch}{1.05}
\resizebox{0.8\linewidth}{!}{
\begin{tabular}{lccc|ccc}
\toprule
\multirow{2}{*}{\textbf{Method}}
& \multicolumn{3}{c|}{\textbf{LP}}
& \multicolumn{3}{c}{\textbf{FT}} \\
\cmidrule(lr){2-4}
\cmidrule(lr){5-7}
& Ch-F1 & AF-F1 & AF-AUROC
& Ch-F1 & AF-F1 & AF-AUROC \\
\midrule

\textbf{SleepXA (Ours)}
& \textbf{73.18}
& \textbf{74.30}
& \underline{91.15}
& \underline{83.35}
& \underline{82.30}
& \underline{98.40} \\

MindMix-style Alignment
& 54.35
& 51.83
& 84.90
& 83.29
& 82.22
& 98.32 \\

Cross-Attention Fusion
& \underline{60.80}
& \underline{61.38}
& \textbf{93.92}
& \textbf{83.73}
& \textbf{83.17}
& \textbf{98.41} \\

Concatenation + MLP Fusion
& 55.97
& 49.79
& 84.49
& 83.22
& 82.29
& \textbf{98.41} \\

DecAlign-style Alignment
& 54.07
& 47.02
& 85.78
& 82.24
& 81.61
& 98.23 \\

CLIP-style Contrastive Alignment
& 57.93
& 54.55
& 90.92
& 82.47
& 81.94
& 98.38 \\

\bottomrule
\end{tabular}
}
\endgroup
\end{table}

\subsubsection{Detailed Ablation of Major Alignment Objectives}
\label{app:major_ablation}

The main text summarizes the effects of the three major SleepXA alignment
mechanisms using representative metrics.
Here, we report the complete LP and FT results for WSC sleep staging,
WSC AHI/OSA prediction, and CinC2017 atrial-fibrillation recognition.
These tasks respectively probe general sleep representation, multisystem
clinical phenotyping, and cross-context single-modality transfer. For each experiment, we independently remove
\textit{condition-aware alignment} (CA),
\textit{stage-conditioned specific preservation} (SP), or
\textit{condition-relative disentanglement} (CD), while keeping the model
architecture, pre-training data, downstream split, and optimization protocol
unchanged.
The complete SleepXA model uses all three mechanisms. Specifically, \textit{w/o CA} retains only instance-level InfoNCE in the
common alignment objective. \textit{w/o SP} removes the specific-preservation objective while
retaining the specific branch and all other training objectives.
\textit{w/o CD} removes the conditional residual-decoupling objective
while retaining all other training objectives.

\paragraph{WSC sleep staging.}
Table~\ref{tab:ablation_wsc_staging} reports the complete sleep-staging
results.
Removing any of the three mechanisms reduces accuracy, Macro-F1, and
Cohen's $\kappa$ under both LP and FT.
Under LP, the complete model reaches an MF1 of 76.09\%, compared with
74.64\%, 74.74\%, and 74.79\% without CA, SP, and CD, respectively.
The same pattern persists after FT, where SleepXA achieves
77.61\% MF1.
The relatively consistent degradation across the three ablations supports
their complementary roles in organizing the learned sleep representation.

\begin{table}[t]
\centering
\caption{
\textbf{Ablation of major objectives on WSC sleep staging.}
Best results within each adaptation setting are shown in \textbf{bold}.
}
\label{tab:ablation_wsc_staging}

\begingroup
\footnotesize
\setlength{\tabcolsep}{3.0pt}
\renewcommand{\arraystretch}{0.92}

\begin{tabularx}{0.6\linewidth}{
@{}cl*{3}{C}@{}
}
\toprule[1.0pt]

\textbf{Mode}
& \textbf{Variant}
& \textbf{Acc. (\%)} $\uparrow$
& \textbf{MF1 (\%)} $\uparrow$
& $\boldsymbol{\kappa}$ $\uparrow$ \\

\midrule


\multirow{4}{*}{\textbf{LP}}
& w/o CA
& 83.90
& 74.64
& 0.749 \\

& w/o SP
& 84.20
& 74.74
& 0.750 \\

& w/o CD
& 83.98
& 74.79
& 0.751 \\

\addlinespace[1.5pt]
& \textbf{SleepXA}
& \textbf{85.16}
& \textbf{76.09}
& \textbf{0.765} \\

\midrule

\multirow{4}{*}{\textbf{FT}}
& w/o CA
& 85.02
& 76.25
& 0.765 \\

& w/o SP
& 85.10
& 76.10
& 0.766 \\

& w/o CD
& 84.87
& 76.05
& 0.763 \\

\addlinespace[1.5pt]
& \textbf{SleepXA}
& \textbf{85.95}
& \textbf{77.61}
& \textbf{0.780} \\

\bottomrule[1.0pt]
\end{tabularx}

\endgroup
\end{table}

\paragraph{WSC AHI estimation and OSA severity classification.}
The effect of the alignment mechanisms becomes more pronounced for the
clinical OSA task (Table~\ref{tab:ablation_wsc_osa}).
Under LP, the complete model achieves the best results across all regression
and classification metrics, including an MAE of 3.441, $R^2$ of 0.864,
and a severity MF1 of 75.05\%.
After FT, removing SP or CD results in larger degradation:
MF1 decreases from 78.46\% to 69.59\% and 71.32\%, respectively,
while the corresponding regression errors also increase.
Both variants also obtain lower severity MF1 after FT than under LP.
One possible explanation is that SP and CD help preserve and organize
modality-specific information relevant to OSA, including SpO$_2$ derived
desaturation patterns and respiratory cues.
Removing these constraints during pre-training may weaken the effective
use of such information during subsequent adaptation, potentially
contributing to the degradation observed after FT.

\begin{table}[t]
\centering
\caption{
\textbf{Ablation of major objectives on WSC AHI estimation and
OSA severity classification.}
Best results within each adaptation setting are shown in \textbf{bold}.
}
\label{tab:ablation_wsc_osa}

\begingroup
\footnotesize
\setlength{\tabcolsep}{2.5pt}
\renewcommand{\arraystretch}{0.92}

\begin{tabularx}{0.8\linewidth}{
@{}cl*{5}{C}@{}
}
\toprule[1.0pt]

\textbf{Mode}
& \textbf{Variant}
& \textbf{MAE} $\downarrow$
& \textbf{RMSE} $\downarrow$
& $\mathbf{R^2}$ $\uparrow$
& \textbf{Acc. (\%)} $\uparrow$
& \textbf{MF1 (\%)} $\uparrow$ \\

\midrule

\multirow{4}{*}{\textbf{LP}}
& w/o CA
& 3.623
& 6.858
& 0.841
& 77.11
& 70.87 \\

& w/o SP
& 3.791
& 6.842
& 0.841
& 78.71
& 72.98 \\

& w/o CD
& 3.756
& 6.921
& 0.838
& 77.51
& 72.70 \\

\addlinespace[1.5pt]
& \textbf{SleepXA}
& \textbf{3.441}
& \textbf{6.341}
& \textbf{0.864}
& \textbf{79.12}
& \textbf{75.05} \\

\midrule


\multirow{4}{*}{\textbf{FT}}
& w/o CA
& 3.361
& 6.430
& 0.860
& 81.53
& 76.07 \\

& w/o SP
& 4.124
& 7.239
& 0.822
& 75.50
& 69.59 \\

& w/o CD
& 3.775
& 7.115
& 0.828
& 78.31
& 71.32 \\

\addlinespace[1.5pt]
& \textbf{SleepXA}
& \textbf{3.120}
& \textbf{6.008}
& \textbf{0.878}
& \textbf{82.73}
& \textbf{78.46} \\

\bottomrule[1.0pt]
\end{tabularx}

\endgroup
\end{table}

\paragraph{CinC2017 AF recognition.}
Table~\ref{tab:ablation_cinc} further evaluates the same mechanisms under
cross-context transfer from nocturnal multimodal PSG to single-lead ECG.
Under LP, the complete SleepXA objective provides a clear advantage in
F1-based recognition, reaching a Ch-F1 of 73.18\% and an AF-F1 of
74.30\%.
Removing CA, SP, or CD reduces AF-F1 to 59.50\%, 57.44\%, and 62.60\%,
respectively.
The highest LP AF-AUROC is obtained without CD, showing that the benefit
is not uniform across all metrics. After FT, however, the differences in F1 become very small.
The variant without CD marginally exceeds the complete model
(82.38\% versus 82.30\% AF-F1), with similarly small differences in
Challenge-F1 and AF-AUROC between these two variants.
On this ECG-only task, task-specific fine-tuning may adapt the ECG
representation sufficiently to compensate for differences introduced
during multimodal pre-training.
This offers a possible explanation for the larger LP gaps and the
similar F1 performance after FT, suggesting that the benefit of these
objectives for AF-F1 is more evident when the pretrained representation
remains frozen.

\begin{table}[t]
\centering
\caption{
\textbf{Ablation of major objectives on CinC2017 AF recognition.}
Ch-F1 and AF-F1 denote the F1 scores for the challenge metric and
atrial fibrillation class, respectively.
Best results within each adaptation setting are shown in \textbf{bold}.
}
\label{tab:ablation_cinc}

\begingroup
\footnotesize
\setlength{\tabcolsep}{3.0pt}
\renewcommand{\arraystretch}{0.92}

\begin{tabularx}{0.82\linewidth}{
@{}cl*{3}{C}@{}
}
\toprule[1.0pt]

\textbf{Mode}
& \textbf{Variant}
& \textbf{Ch-F1 (\%)} $\uparrow$
& \textbf{AF-F1 (\%)} $\uparrow$
& \textbf{AF-AUROC (\%)} $\uparrow$ \\

\midrule


\multirow{4}{*}{\textbf{LP}}
& w/o CA
& 58.79
& 59.50
& 90.83 \\

& w/o SP
& 59.26
& 57.44
& 90.85 \\

& w/o CD
& 62.70
& 62.60
& \textbf{92.37} \\

\addlinespace[1.5pt]
& \textbf{SleepXA}
& \textbf{73.18}
& \textbf{74.30}
& 91.15 \\

\midrule

\multirow{4}{*}{\textbf{FT}}
& w/o CA
& 82.60
& 81.81
& 97.43 \\

& w/o SP
& 82.76
& 82.21
& 97.48 \\

& w/o CD
& \textbf{83.53}
& \textbf{82.38}
& \textbf{98.43} \\

\addlinespace[1.5pt]
& SleepXA
& 83.35
& 82.30
& 98.40 \\

\bottomrule[1.0pt]
\end{tabularx}

\endgroup
\end{table}

Across the three representative tasks, the complete objective shows the
most consistent advantage under LP.
These benefits persist after full fine-tuning for WSC sleep staging and
OSA prediction, whereas the F1 differences in AF recognition largely
diminish.
These findings support the contributions of CA, SP, and CD to learning
transferable representations.

\subsubsection{Alignment with limited sleep stage labels.}
\label{sec:stage_label_effect}
\vspace{-10pt}

\paragraph{Label settings.}
We compare four sleep stage label settings for multimodal alignment.
\emph{SleepXA} uses expert stage labels from SHHS, MESA, and CFS.
\emph{No stage guidance} retains instance contrastive learning,
reconstruction, and other objectives that do not require sleep-stage labels.
\emph{Hybrid pseudo label guidance} retains expert SHHS labels
and uses a SleepXA trained only on SHHS to generate
pseudo labels for MESA and CFS.
The model achieves a pooled staging accuracy of 83.02\%
on the MESA and CFS validation sets.
\emph{Partial label guidance} retains expert SHHS labels
and treats all MESA and CFS recordings as unlabeled.
In this setting, unlabeled epochs are excluded from losses
that require stage labels but still contribute to the other objectives.
We evaluate all settings using LP and FT
on WSC sleep staging, WSC AHI estimation and OSA severity classification,
and CinC2017 AF recognition.
\paragraph{WSC sleep staging.}
Removing stage guidance during pretraining leads to a modest
decline in sleep staging performance
(Table~\ref{tab:stage_guidance_wsc_staging}).
The pretrained encoders may already retain useful staging cues
from EEG and other physiological signals.
Both LP and FT also use stage labels to train the downstream classifier.
These factors may help explain why sleep staging performance
remains relatively high without stage guidance during alignment.
Hybrid pseudo-label guidance and partial label guidance both
improve LP performance over the variant without stage guidance.
These gains suggest that stage guidance makes the pretrained
features more useful for sleep staging.
The FT results remain relatively close across settings.
The gains from the two label settings are larger and more
consistent under LP than under FT.

\begin{table}[t]
\centering
\caption{\textbf{Effect of sleep-stage labels on WSC sleep staging.}
MF1 denotes Macro-F1. Best results within each adaptation setting are
shown in \textbf{bold}. Hybrid and partial variants retain expert SHHS labels.}
\label{tab:stage_guidance_wsc_staging}
\begingroup
\footnotesize
\setlength{\tabcolsep}{3pt}
\renewcommand{\arraystretch}{1.10}
\begin{tabularx}{\linewidth}{@{}cl*{3}{>{\centering\arraybackslash}X}@{}}
\toprule[1pt]
\textbf{Mode} & \textbf{Annotation configuration}
& \textbf{Acc. (\%)} $\uparrow$
& \textbf{MF1 (\%)} $\uparrow$
& $\boldsymbol{\kappa}$ $\uparrow$ \\
\midrule
\multirow{4}{*}{\textbf{LP}}
& No stage guidance & 83.71 & 74.92 & 0.751 \\
& Hybrid pseudo label guidance & 84.69 & 75.89 & 0.762 \\
& Partial label guidance & 84.23 & 75.57 & 0.760 \\
& \textbf{SleepXA (full labels)}
& \textbf{85.16} & \textbf{76.09} & \textbf{0.765} \\
\midrule
\multirow{4}{*}{\textbf{FT}}
& No stage guidance & 85.40 & 76.35 & 0.770 \\
& Hybrid pseudo-label guidance & 85.51 & 77.19 & 0.774 \\
& Partial label guidance & 84.90 & 76.20 & 0.765 \\
& \textbf{SleepXA (full labels)}
& \textbf{85.95} & \textbf{77.61} & \textbf{0.780} \\
\bottomrule[1pt]
\end{tabularx}
\endgroup
\end{table}

\vspace{-5pt}
\paragraph{WSC AHI estimation and OSA severity classification.}
Removing stage guidance weakens AHI estimation and OSA severity
classification over the whole night
(Table~\ref{tab:stage_guidance_wsc_osa}).
The drop in classification accuracy is larger than that observed
for sleep staging.
Hybrid pseudo label guidance and partial label guidance both
improve AHI estimation and classification accuracy.
Their gains in LP MF1 are modest.
The benefits remain after FT, although fully annotated SleepXA
still performs best.
These results support the value of stage guidance for overnight
clinical assessment. Both label settings recover part of this
benefit with fewer expert annotations.

\begin{table}[t]
\centering
\caption{\textbf{Effect of sleep-stage labels on WSC AHI estimation
and OSA severity classification.}
Acc. and MF1 evaluate OSA severity classification.
Best results within each adaptation setting are shown in \textbf{bold}.}
\label{tab:stage_guidance_wsc_osa}
\begingroup
\footnotesize
\setlength{\tabcolsep}{2.5pt}
\renewcommand{\arraystretch}{1.10}
\begin{tabularx}{\linewidth}{@{}cl*{5}{>{\centering\arraybackslash}X}@{}}
\toprule[1pt]
\textbf{Mode} & \textbf{Annotation configuration}
& \textbf{MAE} $\downarrow$
& \textbf{RMSE} $\downarrow$
& $\mathbf{R^2}$ $\uparrow$
& \textbf{Acc. (\%)} $\uparrow$
& \textbf{MF1 (\%)} $\uparrow$ \\
\midrule
\multirow{4}{*}{\textbf{LP}}
& No stage guidance & 3.655 & 6.584 & 0.853 & 73.41 & 73.68 \\
& Hybrid pseudo label guidance & 3.489 & 6.356 & 0.863 & 78.31 & 73.88 \\
& Partial label guidance & 3.508 & 6.556 & 0.854 & 77.51 & 73.73 \\
& \textbf{SleepXA (full labels)}
& \textbf{3.441} & \textbf{6.341} & \textbf{0.864}
& \textbf{79.12} & \textbf{75.05} \\
\midrule
\multirow{4}{*}{\textbf{FT}}
& No stage guidance & 3.621 & 6.611 & 0.852 & 78.72 & 75.19 \\
& Hybrid pseudo label guidance & 3.280 & 6.492 & 0.857 & 80.72 & 77.83 \\
& Partial label guidance & 3.410 & 6.508 & 0.856 & 80.31 & 77.31 \\
& \textbf{SleepXA (full labels)}
& \textbf{3.120} & \textbf{6.008} & \textbf{0.878}
& \textbf{82.73} & \textbf{78.46} \\
\bottomrule[1pt]
\end{tabularx}
\endgroup
\end{table}
\vspace{-5pt}
\paragraph{CinC2017 AF recognition.}
The CinC2017 task tests ECG transfer outside the sleep setting
(Table~\ref{tab:stage_guidance_cinc}).
LP performance drops sharply when stage guidance is removed.
This suggests that stage guidance also benefits ECG features
used for tasks beyond sleep staging.
Hybrid pseudo-label guidance and partial label guidance both
bring LP performance close to that of fully annotated SleepXA.
During FT, labels from the AF task guide updates to both the
encoder and the classifier. This may help compensate for
differences in the pretrained features and explain the smaller
performance gaps.
The value of stage guidance is therefore clearest when the
pretrained encoders remain frozen.

\begin{table}[t]
\centering
\caption{\textbf{Effect of sleep-stage labels on CinC2017 AF recognition.}
All scores, including AF-AUROC, are reported as percentages.
Best results within each adaptation setting are shown in \textbf{bold}.}
\label{tab:stage_guidance_cinc}
\begingroup
\footnotesize
\setlength{\tabcolsep}{3pt}
\renewcommand{\arraystretch}{1.10}
\begin{tabularx}{\linewidth}{@{}cl*{3}{>{\centering\arraybackslash}X}@{}}
\toprule[1pt]
\textbf{Mode} & \textbf{Annotation configuration}
& \textbf{Challenge-F1} $\uparrow$
& \textbf{AF-F1} $\uparrow$
& \textbf{AF-AUROC} $\uparrow$ \\
\midrule
\multirow{4}{*}{\textbf{LP}}
& No stage guidance & 59.51 & 60.24 & 87.40 \\
& Hybrid pseudo label guidance & 72.91 & 73.35 & 89.60 \\
& Partial label guidance & 72.60 & 72.12 & 88.60 \\
& \textbf{SleepXA (full labels)}
& \textbf{73.18} & \textbf{74.30} & \textbf{91.15} \\
\midrule
\multirow{4}{*}{\textbf{FT}}
& No stage guidance & 82.69 & 81.85 & 98.30 \\
& Hybrid pseudo label guidance & 82.71 & 81.56 & 98.20 \\
& Partial label guidance & 82.39 & 81.41 & 98.10 \\
& \textbf{SleepXA (full labels)}
& \textbf{83.35} & \textbf{82.30} & \textbf{98.40} \\
\bottomrule[1pt]
\end{tabularx}
\endgroup
\end{table}

These results indicate that SleepXA’s sleep stage-based objectives
help facilitate transfer to downstream tasks. The specific effectiveness depends
on the task and the adaptation settings. When human expert annotation labels are limited, hybrid pseudo label guidance and partial label guidance offer
a viable alternative.

\subsection{Robustness under Restricted Modalities and Supervision}
\label{app:robustness_analysis}

We further evaluate SleepXA under two practical constraints that are not
fully explored in the main text:
restricted modality availability and limited downstream supervision.
These experiments examine whether the learned representation remains useful
when the downstream sensor configuration differs from the full PSG setting
or when only a small number of labeled subjects are available.

\subsubsection{Robustness to Restricted Modality Availability}
\label{app:modality_robustness}

\begin{figure}[t]
    \centering
    \includegraphics[
        width=\columnwidth
    ]{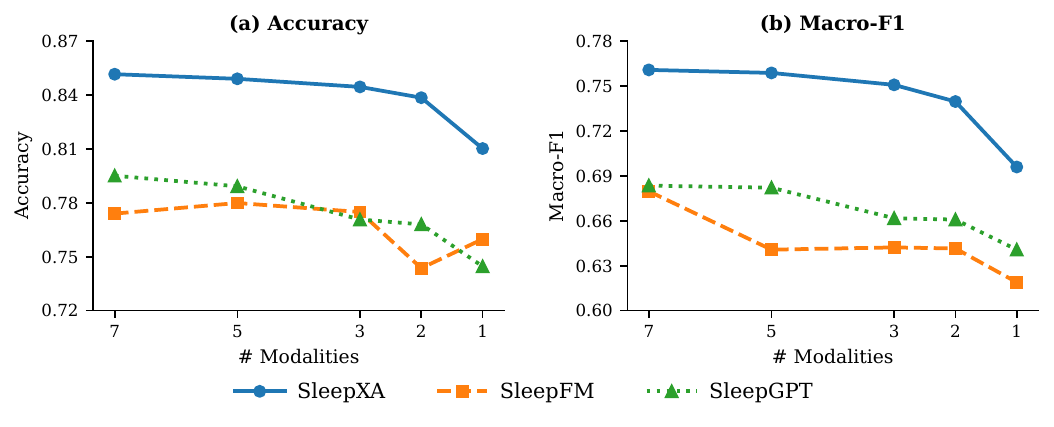}
    \vspace{-25pt}
    \caption{
    \textbf{Robustness to restricted modality availability.}
    WSC sleep staging under LP with progressively reduced
    physiological inputs, from the full seven-modality PSG configuration
    to EEG-only evaluation.
  }
    \label{fig:modality_ablation}
    \vspace{-6pt}
\end{figure}

We examine the effect of reducing the number of available physiological
modalities at downstream evaluation.
Experiments are conducted on WSC sleep staging under LP so that
the pretrained backbone remains fixed, and the comparison primarily reflects
the quality of the learned representations.
Starting from the full seven-modality PSG configuration
(EEG, EOG, EMG, ECG, airflow, ABD, and SpO$_2$),
we progressively reduce the available inputs down to an EEG-only setting.
For comparison, we include SleepFM and SleepGPT, which can be evaluated under
the same modality configurations as SleepXA, enabling a controlled comparison
as sensor availability is reduced. As shown in Fig.~\ref{fig:modality_ablation}, SleepXA maintains strong
performance across the evaluated configurations, with gradual degradation as
modalities are removed.
The representation remains effective even under substantially restricted
sensor availability, including the EEG-only setting.
These results indicate that SleepXA can exploit complementary multimodal
information when available without requiring a fixed full-PSG configuration
at downstream evaluation.

\subsubsection{Pre-training Scale under Limited Downstream Supervision}
\label{app:fewshot_analysis}
We next investigate how the scale and cohort diversity of sleep-domain
pre-training interact with the amount of labeled downstream data.
All experiments use LP. For each downstream task, the original validation and test partitions are
kept fixed while the number of labeled training subjects is varied.
We consider five labeling ratios:
5\%, 10\%, 25\%, 50\%, and 100\%.
For WSC, these correspond to 45, 90, 225, 450, and 899 labeled training
subjects, respectively, with 111 subjects each in the validation and test
sets.
Each experiment is repeated using five downstream random seeds
$\{40,41,42,43,44\}$.
For each seed, subjects are randomly ordered once, and progressively larger
training subsets are constructed from the same ordering, ensuring that the
smaller labeled subsets are nested within the larger ones.

We compare three pre-training configurations:
\textit{Single-domain} uses the SHHS pre-training cohort; \textit{Multi-domain (Matched)} uses SHHS, MESA, and CFS while matching the
total number of pre-training participants to the single-domain setting; and
\textit{Multi-domain (Full)} uses all available SHHS, MESA, and CFS
pre-training data.
The pretrained checkpoint is fixed within each configuration, and downstream
seeds affect only subject subsampling and prediction-head optimization.
Results are summarized as mean deviation across the five
downstream seeds.
For SEED-VIG, performance is first averaged across the five subject-level
folds within each seed and then aggregated across seeds.

\begin{figure*}[t]
    \centering
    \includegraphics[
        width=\textwidth
    ]{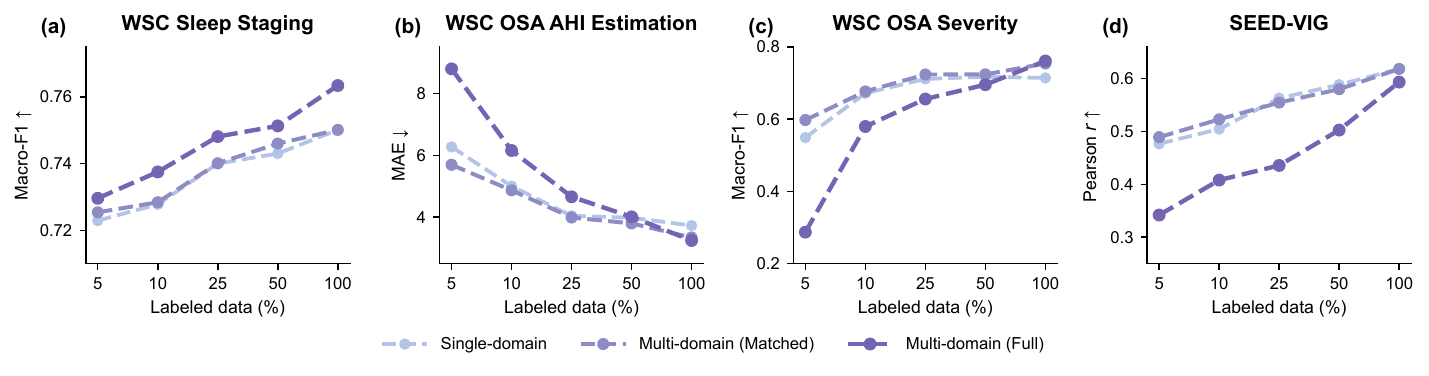}
    \vspace{-15pt}
    \caption{
    \textbf{Effect of pre-training diversity and scale under limited downstream supervision}
    LP performance using 5\%, 10\%, 25\%, 50\%, and 100\% of the
    labeled downstream training subjects under single-domain,
    matched multi-domain, and full multi-domain pre-training.
    (a) WSC sleep staging, evaluated by Macro-F1;
    (b) WSC AHI estimation, evaluated by MAE;
    (c) WSC OSA severity classification, evaluated by Macro-F1; and
    (d) SEED-VIG vigilance estimation, evaluated by Pearson's $r$.
    }
    \label{fig:fewshot}
    \vspace{-6pt}
\end{figure*}

Figure~\ref{fig:fewshot} reveals a task-dependent effect of increasing the
scale and diversity of nocturnal PSG pre-training.
For WSC sleep staging, full multi-domain pre-training consistently provides
the strongest transfer across labeling ratios, with the benefit already
visible in the low-label regime.
For AHI estimation and OSA severity classification, the effect is more
dependent on downstream supervision: the matched multi-domain configuration
is often stronger when labels are scarce, whereas the full multi-domain
representation becomes more competitive as supervision increases.
A different trend is observed for SEED-VIG, where increasing sleep-domain
pre-training scale does not produce monotonic improvements. Together, these results suggest that scaling heterogeneous PSG pre-training
most reliably benefits downstream tasks that remain closely aligned with
nocturnal sleep physiology.
For more distant clinical or daytime transfer settings, the usefulness of
additional pre-training scale depends more strongly on the amount of
task-specific supervision and the degree of domain shift.

\subsection{Additional Methodological Details}
\label{app:common_alignment}
\paragraph{Instance and sleep-stage alignment.}
The two terms in Eq.~\ref{eq:sleep_guided_alignment} implement
contrastive alignment at the instance and sleep-stage levels.
For $\mathcal L_{\mathrm{inst}}$, synchronized representations from
two available modalities of the same epoch form a positive pair,
while eligible epochs from other participants in the same cohort
serve as negatives. The objective is computed bidirectionally
between each valid modality pair.

For $\mathcal L_{\mathrm{stage}}$, we maintain learnable
modality-invariant prototypes ${\mathbf p_y}$ for the five
sleep stages. Each common representation is contrasted against
these prototypes, with the prototype of its sleep stage as the
positive. In the reverse direction, each prototype is contrasted
against the corresponding batch stage centroids computed within
each cohort and modality. These two directions are combined to
obtain $\mathcal L_{\mathrm{stage}}$.

\begin{equation}
\mathcal L_{\mathrm{inst}}=-\mathbb E_{i,m,n}
\log
\frac{
\exp\!\left(
\operatorname{sim}(\mathbf c_i^m,\mathbf c_i^n)/\tau
\right)
}{
\exp\!\left(
\operatorname{sim}(\mathbf c_i^m,\mathbf c_i^n)/\tau
\right)
+
\sum_{j\in\mathcal N_i^n}
\exp\!\left(
\operatorname{sim}(\mathbf c_i^m,\mathbf c_j^n)/\tau
\right)
}.
\label{eq:app_instance}
\end{equation}
where $\mathcal N_i^n$ denotes eligible negatives from other
participants in the same cohort. The objective is applied
bidirectionally across valid modality pairs.

\paragraph{Structural alignment.}

To implement Eq.~\ref{eq:conditional_alignment}, we group common
representations by cohort and sleep stage, and compare each modality
pair using epochs in which both modalities are available. For each
valid group, $D_{\mathrm{struct}}$ combines the cosine distance
between paired representations with differences in their feature-wise
means and standard deviations. All comparisons are computed on
normalized common representations and aggregated across valid
modality pairs, stages, and cohorts.

For a valid cohort--stage group $\mathcal G_{d,y}^{m,n}$, we define
\begin{equation}
\begin{aligned}
D_{\mathrm{pair}}
&=
\mathbb E_{i\in\mathcal G_{d,y}^{m,n}}
\left[
1-\cos(\mathbf u_i^m,\mathbf u_i^n)
\right],\
D_{\mathrm{struct}}
&=
\frac{1}{2}
\left(
D_{\mathrm{pair}}+D_{\mathrm{stat}}
\right),
\end{aligned}
\label{eq:app_structural}
\end{equation}
where $\mathbf u_i^m$ denotes the parameter-free LayerNorm of
$\mathbf c_i^m$, and $D_{\mathrm{stat}}$ measures discrepancies
in the feature-wise means and standard deviations between the
two conditional representation distributions.

\paragraph{Modality-specific preservation.}
For each cohort and sleep stage, we consider the modalities observed
in that group and normalize their specific representations and
learnable prototypes. In the representation-to-prototype direction,
each $\mathbf s_i^m$ is contrasted against the prototypes of the
observed modalities for stage $y_i$, with $\mathbf q_{y_i}^m$ as its
positive target. In the prototype-to-centroid direction, we first
compute a centroid for each observed modality within the cohort-stage
group. Each prototype is then contrasted against these centroids,
with the centroid of its own modality as the positive target.
Both directions use temperature-scaled cross-entropy and are applied
only when at least two modalities are observed in the group. The
resulting terms are aggregated across valid stages and cohorts. We set $\alpha_{\mathrm{stage}}=0.5$,
$\lambda_{\mathrm{spec}}=\lambda_{\mathrm{dis}}=0.1$,
and $\lambda_{\mathrm{rec}}=1$ throughout.

\end{document}